\pdfoutput=1

\documentclass[11pt]{article}

\usepackage[final]{acl}

\usepackage{times}
\usepackage{latexsym}

\usepackage[T1]{fontenc}

\usepackage[utf8]{inputenc}

\usepackage{microtype}

\usepackage{inconsolata}

\usepackage{graphicx}
\usepackage{booktabs}
\usepackage{multirow}
\usepackage{amsmath}
\usepackage{makecell}
\usepackage{amssymb}
\usepackage{pifont}
\usepackage{xcolor}
\usepackage[most]{tcolorbox}
\usepackage{float}
\usepackage{placeins}
\renewcommand{\textcolor}[2]{#2}
\title{\textit{MapCoder-Lite}: Distilling Multi-Agent Coding into a Single Small LLM}

\author{
  Woongkyu Lee$^{1}$ \quad Junhee Cho$^{2}$ \quad Jungwook Choi$^{1}$\thanks{Corresponding author.} \\
  $^1$Hanyang University \quad 
  $^2$Samsung SDS \\
  \texttt{\{lwghanyang, choij\}@hanyang.ac.kr} \\
  \texttt{junhee.cho@samsung.com}
}

\begin{document}
\maketitle
\begin{abstract}

Large language models (LLMs) have advanced code generation from single-function tasks to \textit{competitive-programming} problems, but existing multi-agent solutions either rely on costly large-scale ($> 30 B$) models or collapse when downsized to small open-source models. We present \emph{MapCoder-Lite}, a framework for distilling the complex reasoning of large, multi-agent coding systems into a single 7B model. Our contribution is a novel, three-pillar methodology that synergistically generates, refines, and encodes multi-agent knowledge:
(i) \emph{pass-based trajectory distillation} from strong LLMs fixes format fragility in retrieval and reduces failures in debugging, (ii) \emph{supervisor-guided correction} with global feedback strengthens planning and coding agents, and (iii) \emph{agent-wise LoRA fine-tuning} delivers memory-efficient specialisation.
Comprehensive evaluation on xCodeEval, APPS, and CodeContests shows that MapCoder-Lite more than doubles xCodeEval accuracy (13.2\% → 28.3\%), eliminates all format failures, while reducing GPU memory and token-generation time by $4\times$ compared to a 32B model. It also achieves over 10\% gains on simpler coding benchmarks, demonstrating broad improvements beyond competitive programming.
These results demonstrate that careful agent-wise fine-tuning unleashes high-quality multi-agent coding on a small language model. Our code is publicly available at \url{https://github.com/aiha-lab/MapCoder-Lite}.

\end{abstract}

\section{Introduction}

LLMs have revolutionized code synthesis, achieving near-perfect accuracy on function-level tasks like HumanEval~\cite{chen2021evaluating} and MBPP~\cite{austin2021programsynthesis}. Research has now shifted to \textit{competitive programming}, which demands efficient algorithms, robust implementation, and resilience against hidden test cases. \textcolor{blue}{Benchmarks like CodeElo~\cite{quan2025codeelo} highlight significant challenges that have spurred the development of massive models, such as OpenAI's o3~\cite{openai2025competitive} and the 671B-parameter DeepSeek-V3~\cite{deepseekai2025deepseekv3}. These models report strong performance on Codeforces and related benchmarks despite their computational costs and proprietary nature.}



\begin{figure*}[t]
\centering
\includegraphics[width=\textwidth]{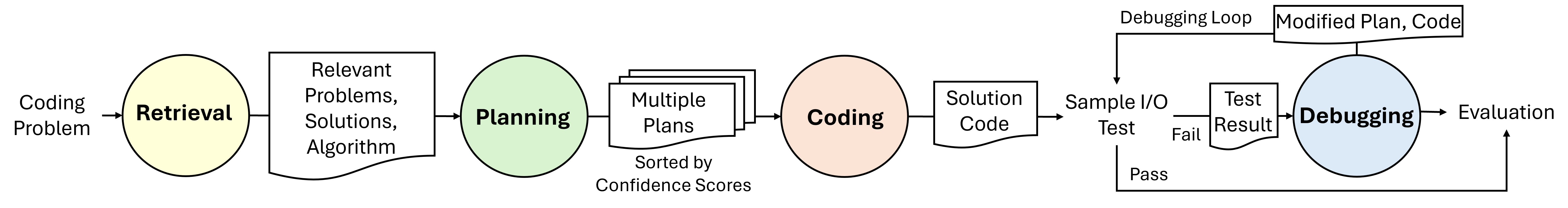}
\caption{Overview of the MapCoder system. Given a natural language problem, the retrieval agent fetches relevant algorithmic knowledge, followed by the planning agent generating a solution plan. The coding agent implements the plan, and the debugging agent iteratively refines the code based on test outcomes.}
\label{fig:mapcoder-overview}
\end{figure*}

Competitive programming involves algorithmically complex problem solving under strict constraints, making it a difficult setting for language models. Single-agent prompting, where one model handles the entire problem-solving process, often falls short~\cite{wei2023chainofthought, jiang2024selfplanning, yasunaga2024analogical}. To overcome this, recent work has explored \textit{multi-agent code-generation} frameworks that split the task into stages and assign each to a dedicated agent, improving end-to-end performance through role specialization~\cite{huang2024agentcoder, hong2024metagpt, islam-etal-2024-mapcoder, islam2025codesim}. MapCoder~\cite{islam-etal-2024-mapcoder} (Fig.~\ref{fig:mapcoder-overview}) exemplifies this approach, coordinating specialized agents throughout the pipeline.

The effectiveness of multi-agent frameworks typically relies on large-scale ($>30B$) open models~\cite{openai2024gpt4technicalreport, comanici2025gemini25pushingfrontier}, which possess the capacity to perform a wide range of specialized roles required across the multi-agent pipeline. Due to their multi-step nature, these frameworks incur significantly higher token usage and API calls than single-agent setups, resulting in increased latency and computational cost.

Their impracticality in resource-constrained settings naturally motivates small-model multi-agent solutions, which align well with the growing industry trend toward on-device AI~\cite{gunter2024appleintelligencefoundationlanguage, zhang2023appagentmultimodalagentssmartphone}. \textcolor{blue}{In practice, however, deploying small language models (SLMs) under 10B parameters—even those with strong coding abilities~\cite{qwen2025qwen25technicalreport}—often yields limited accuracy gains compared to single-model direct prompting.} This performance drop stems from SLMs’ difficulty in following the structured formats (e.g., XML) required for multi-agent communication~\cite{xia2024fofobenchmarkevaluatellms} and their limited capacity to support the complex reasoning needed across agent roles, leading to failures in retrieval, planning, or debugging.


To make SLM-based multi-agent systems effective, fine-tuning becomes a necessary step. However, this approach poses three key obstacles. First, existing code datasets do not provide intermediate artefacts aligned with the roles in a multi-agent pipeline. Second, fine-tuning each agent independently fails to account for inter-agent dependencies. As illustrated in Fig.~\ref{fig:mapcoder-overview}, later stages rely on the outputs of earlier agents, so errors in one stage can propagate and mislead downstream components. Third, training separate models for each agent increases GPU memory consumption at inference time, undermining the efficiency benefits of using small models in the first place.

We address these challenges with \emph{MapCoder-Lite}, the first multi-agent coding framework that drives a single 7B backbone--extended only by lightweight, role-specific adapters--to performance near that of 32B systems. \emph{MapCoder-Lite} is built on three components:
\begin{itemize}
    \item \textbf{Pass-based trajectory distillation from strong LLMs (Sect.~\ref{subsec:method1})}: 
    To address the lack of role-specific training data, we collect trajectories from strong LLMs. However, fine-tuning on outputs that are only locally valid often fails to yield correct final solutions. We overcome this by implementing \textit{pass-based filtering} that exclusively retains trajectories whose final code passes all unit tests.
    \item \textbf{Supervisor-aided cross-agent refinement (Sec.~\ref{subsec:method2})}: We employ a \textit{supervisor} model that analyzes the full trajectory generated by the small model to detect cross-agent failure patterns and regenerate the responsible agent’s output, guiding the model toward global success and helping bridge the capacity gap.
    \item \textbf{Memory-efficient LoRA specialization (Sec.~\ref{subsec:method3})}: We show that in complex multi-agent settings, LoRA~\cite{hu2021lora} achieves better accuracy and efficiency than full fine-tuning. This enables all agents to share a frozen Qwen2.5-7B backbone with lightweight, role-specific adapters, adding under 3\% extra parameters.
\end{itemize}

We conducted a comprehensive evaluation on three representative competitive-programming suites—xCodeEval~\cite{khan-etal-2024-xcodeeval}, APPS~\cite{hendrycks2021apps}, and CodeContests~\cite{doi:10.1126/science.abq1158_codecontest}—and found that \emph{MapCoder-Lite} leverages trajectory distillation, supervisor-guided cross-agent refinement, and rank-32 LoRA specialisation to boost xCodeEval accuracy from 13.2\% to 28.3\%, eliminates every XML-schema failure, all while cutting GPU memory and token generation time(time per output token) by $4\times$. These results demonstrate that careful agent-wise fine-tuning can unlock high-quality multi-agent code generation on small language models.

\section{Related Work}

\begin{figure*}[t]
\centering
\includegraphics[width=\textwidth]{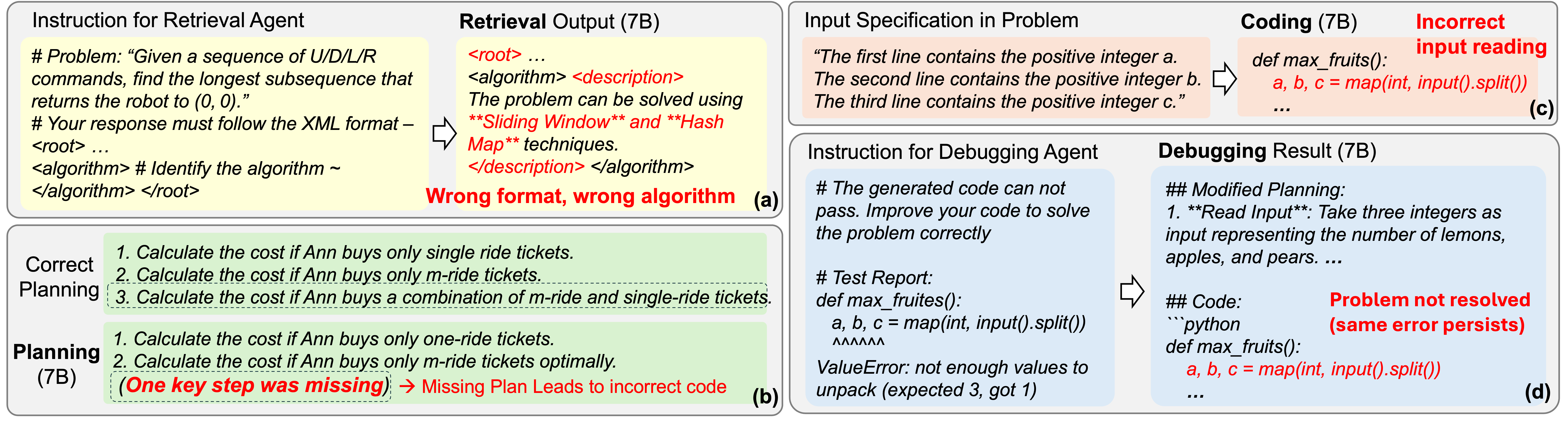}
\caption{
Representative failure cases of the 7B-scale model across all agents. (a) Retrieval: invalid XML format and incorrect algorithm. (b) Planning: missing key step. (c) Coding: misinterpreted input specification. (d) Debugging: persistent unresolved error. Detailed illustrations are provided in Appendix~\ref{sec:improvements}.}
\label{fig:failure-cases}
\end{figure*}

\subsection{LLMs for Competitive Programming}
LLMs have shown strong performance on \emph{function-level} code generation tasks~\cite{hui2024qwen25coder,guo2024deepseekcoder,rozière2024codellama}, achieving near-perfect scores on HumanEval~\cite{chen2021evaluating} and MBPP~\cite{austin2021programsynthesis}.
Recent work shifts to the harder setting of \emph{competitive programming}, which requires generating full, efficient programs that pass hidden tests.
Benchmarks like APPS~\cite{hendrycks2021apps}, CodeContests~\cite{doi:10.1126/science.abq1158_codecontest}, and xCodeEval~\cite{khan-etal-2024-xcodeeval} capture this challenge and now define the state of the art~\cite{openai2025competitive,deepseekai2025deepseekr1}, motivating the multi-agent systems that follow.

\subsection{Multi-Agent Code Generation}
Recent work has proposed \textit{multi-agent pipelines} to better handle the linguistic complexity and algorithmic subtlety of competitive programming, which often cause single prompts to miss edge cases or violate constraints~\cite{huang2024agentcoder, pan2025codecor}. Among these, \textit{MapCoder}\cite{islam-etal-2024-mapcoder} adopts a four-stage pipeline (Fig.\ref{fig:mapcoder-overview}) that performs well on APPS and CodeContests. A \emph{retrieval} agent first selects an algorithm from a private corpus and returns a schema-constrained XML snippet. A \emph{planning} agent then expands this into step-wise plans with confidence scores. The top plan is passed to a \emph{coding} agent, which generates the full source code. Finally, a \emph{debugging} agent tests and patches the code until it passes all unit tests, backtracking to alternative plans if needed. This modular design improves accuracy via role specialization, but also demands broad capability from the underlying LLM across all stages.

\subsection{Task-Specific Fine-Tuning}
MapCoder uses strong LLMs at each stage, trading efficiency for accuracy. A natural question is whether a single small model (<10B), fine-tuned per role, can achieve similar results. While prior work has shown the value of fine-tuning in multi-agent~\cite{zhao2025sirius, liang2025cmat, shen-etal-2024-small} and code tasks~\cite{fan2025fait, tsai2024codeless, 10.1145/3695993, NEURIPS2024_ledex}, no study has systematically explored role-aligned fine-tuning for small LLMs in competitive programming pipelines. To our knowledge, however, no work has comprehensively studied how far role-aligned fine-tuning can push a small LLM inside a multi-agent pipeline for competitive programming.

\section{Analysis of Multi-Agent Limitations}


\subsection{High Cost with Large Models}  
We evaluated MapCoder using the Qwen2.5-32B-Instruct model on the CodeContests benchmark, which comprises 165 competitive programming problems. 
The system required 27.53 hours of runtime, processed approximately 5.08 million input tokens and 1.60 million output tokens, and made 3,095 API calls. This substantial resource usage stems from MapCoder’s multi-agent design involving four distinct agents and multiple iterations for planning and debugging. These results highlight the heavy runtime and memory burden of using large-scale language models throughout the pipeline. \textcolor{blue}{We therefore hypothesize that replacing each agent with a fine-tuned SLM can substantially reduce token-generation time and GPU memory usage, even under comparable token and API call counts.}


\subsection{Failure Cases of Small Models}


\textbf{Format Following Failures.} Multi-agent systems often rely on structured outputs in predefined formats~\cite{yang2025docagentmultiagentautomatedcode, tang2025autoagentfullyautomatedzerocodeframework}. In MapCoder, for example, retrieval and planning agents are required to produce XML-formatted responses such as <root>, <algorithm>, and <confidence> for downstream parsing. However, small models frequently violate these schemas—omitting or misplacing tags—which disrupts subsequent processing and halts pipeline execution (Fig.~\ref{fig:failure-cases}a). This issue is amplified by the weaker format-following ability of open-source SLMs compared to proprietary LLMs~\cite{xia2024fofobenchmarkevaluatellms}, underscoring the importance of strict structural adherence in multi-agent workflows.

\textbf{Low Role Performance.} Small models often struggle with role-specific tasks due to limited capacity. Fig.~\ref{fig:failure-cases} illustrates representative failures across agents:
\textcolor{blue}{
(a) the retrieval agent produces incorrect or misleading algorithm descriptions, corrupting shared context for all downstream stages;
(b) the planning agent outputs superficially correct but incomplete plans, often missing subtle edge cases;
(c) the coding agent introduces logical or I/O errors even when given valid plans, yielding incorrect or unexecutable code; and
(d) the debugging agent fails to detect or fix simple bugs, resulting in repeated ineffective patches.
Together, these failures indicate that, without stronger supervision or greater capacity, individual agents act as bottlenecks that undermine the reliability of the entire pipeline.}

\section{Challenges}
\label{sec:challenge}


\textcolor{blue}{The multi-agent approach provides a parameter-free mechanism for orchestrating specialized roles via prompting, yet SLMs consistently underperform in this zero-shot setting. This suggests that prompt-based assignment alone is insufficient for SLMs, necessitating a shift toward parameter-driven optimization~\cite{du2025surveyoptimizationlargelanguage}. However, effectively fine-tuning SLMs within a multi-agent framework presents several non-trivial challenges.}

\subsection{Lack of Role-Specific Training Data.}
Multi-agent fine-tuning requires high-quality intermediate supervision tailored to each agent’s role. Existing approaches rely either on reconstructing intermediate signals from public datasets~\cite{shen-etal-2024-small} or on self-collection using SLM-generated outputs~\cite{liang2025cmat}. However, public code benchmarks are not designed for multi-agent pipelines and lack role-aligned artefacts, while self-collected data from small models is often malformed or incomplete due to limited capacity and task complexity. This underscores the need to generate new, role-specific supervision tailored to multi-agent training.

\begin{figure*}[t]
\centering
\includegraphics[width=\textwidth]{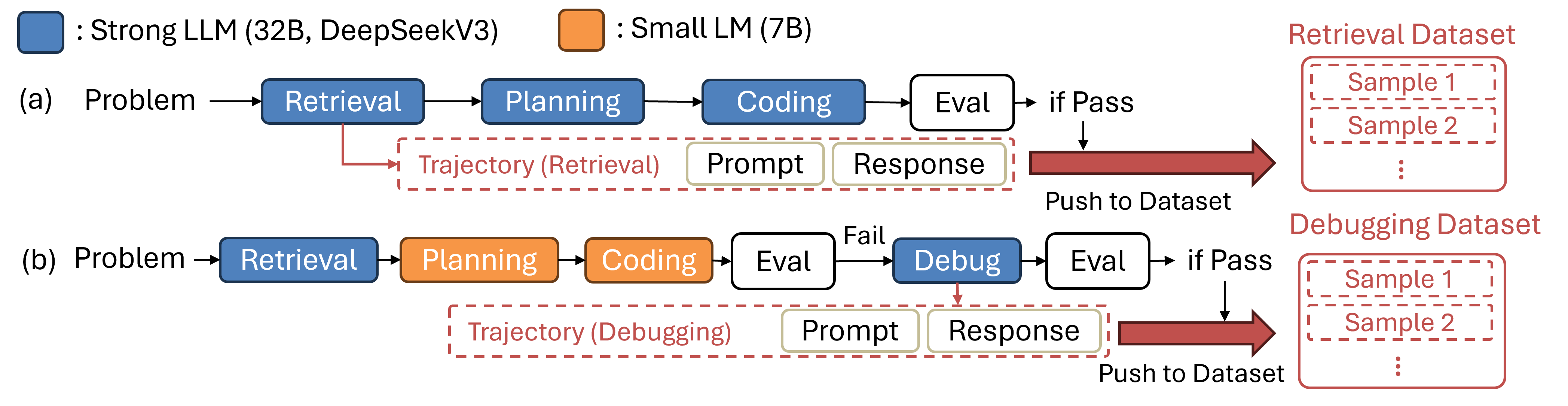}
\caption{Illustration of trajectory construction for retrieval and debugging datasets. (a) When a strong LLM generates a complete solution that passes unit tests, the retrieval agent’s input-output pair is extracted as a training example. (b) To collect debugging data, a 7B model is used for planning and coding, and the strong LLM is used for debugging when the initial code fails. If the revised output passes, the debugging trajectory is added to the dataset.}
\label{fig:trajectory}
\end{figure*}

\subsection{Limited Global Awareness.}
In multi-agent workflows, information flows sequentially, making end-to-end success reliant on the coherence of intermediate outputs across stages. Yet agents trained in isolation lack awareness of these dependencies, leading to local errors that propagate and compromise the final outcome. For instance, when the final code fails, it is often unclear whether the root cause lies in a flawed plan from the planning agent or in an incorrect implementation by the coding agent, even if each agent appears to perform its role plausibly. Such failures show that success depends not only on individual competence but also on aligned interactions. Without global awareness, even well-tuned agents may fall short.

\subsection{Inefficiency of Full Model Fine-Tuning}
Previous multi-agent fine-tuning approaches typically rely on OpenAI’s fine-tuning API~\cite{zhao2025sirius} or full-parameter tuning for each agent~\cite{shen-etal-2024-small, zeng2023agenttuningenablinggeneralizedagent}. However, full fine-tuning scales memory usage linearly with the number of agents, negating one of the primary advantages of SLMs—their efficiency in memory and deployment. Furthermore, full fine-tuning does not necessarily guarantee superior performance over parameter-efficient methods such as LoRA~\cite{hu2021lora}.


\section{Methodology}

To overcome the limitations of applying multi-agent code generation to SLMs, we propose a \textit{role-aligned supervised fine-tuning} pipeline that equips a single 7B backbone with specialized behavior for four agents via lightweight LoRA adapters. This design directly addresses the three main challenges identified in our analysis:

\begin{itemize}
\item \textbf{Lack of role-specific data.}
To compensate for the absence of intermediate supervision, we distill high-quality trajectories from strong LLMs, using execution results to retain only clean, agent-specific samples (Sec.~\ref{subsec:method1}).
\item \textbf{Limited global awareness.}
To maintain cross-agent consistency, we introduce a \textit{supervisor-guided refinement} mechanism that detects failures and regenerates only the faulty component, yielding coherent, context-aware training data (Sec.~\ref{subsec:method2}).
\item \textbf{Inefficiency of full-model fine-tuning.}
We address parameter and memory overhead by applying rank-32 LoRA adapters to a shared frozen 7B backbone, enabling agent-wise specialization with minimal additional cost (Sec.~\ref{subsec:method3}).
\end{itemize}

Together, these techniques allow us to retain the advantages of the multi-agent approach while making it viable for deployment on open-source, resource-efficient language models.


\subsection{Strong LLM for Retrieval and Debugging}
\label{subsec:method1}

Fig.~\ref{fig:trajectory} shows the proposed trajectory construction method. Our data pipeline begins by asking \textit{strong LLMs}, \textcolor{blue}{namely Qwen2.5-32B-Instruct and DeepSeek-V3, to solve each coding task while explicitly printing intermediate artefacts (i.e., \textit{trajectories}) produced by each MapCoder role.}

To build reliable training data, we collect trajectories from strong LLMs and keep only those whose final code passes all unit tests. Unlike rejection sampling based solely on local validity at the single-agent level~\cite{zelikman2022star, zeng2023agenttuningenablinggeneralizedagent}, our \emph{pass-based filtering} ensures end-to-end success across all roles. This focuses fine-tuning on trajectories verified through full execution, yielding accuracy gains over locally valid samples (Table~\ref{tab:supervisor-ablation}, row 2 vs. row 3).


\begin{table}[t]
\centering
\resizebox{\linewidth}{!}{%
\begin{tabular}{cc|cc|c}
\toprule
\multicolumn{2}{c|}{\textbf{Retrieval}} & \multicolumn{2}{c|}{\textbf{Planning / Coding}} & \textbf{xCodeEval} \\
\cmidrule(lr){1-2} \cmidrule(lr){3-4} \cmidrule(l){5-5}
\textbf{Dataset} & \textbf{Filtering} & \textbf{Dataset} & \textbf{Filtering} & \textbf{Accuracy (\%)} \\
\midrule
-- & -- & -- & -- & 11.32 \\
Strong & -- & -- & -- & 16.04 \\
Strong & Format  & -- & -- & 16.98 \\
Strong & Pass  & Strong     & Pass & 18.87 \\
\textbf{Strong} & \textbf{Pass}  & \textbf{Supervisor}  & \textbf{Pass} & \textbf{22.64} \\
\bottomrule
\end{tabular}
}
\caption{Ablation study on data source and filtering methods for training retrieval, planning, and coding agents on xCodeEval. “Format” filtering keeps samples with valid single-agent outputs, while “Pass” retains only those where the final program passes all unit tests. “Strong” denotes data from a large LLM; “Supervisor” indicates trajectories refined after failure analysis.}
\label{tab:supervisor-ablation}
\end{table}

\begin{figure}[t]
\centering
\includegraphics[width=\linewidth]{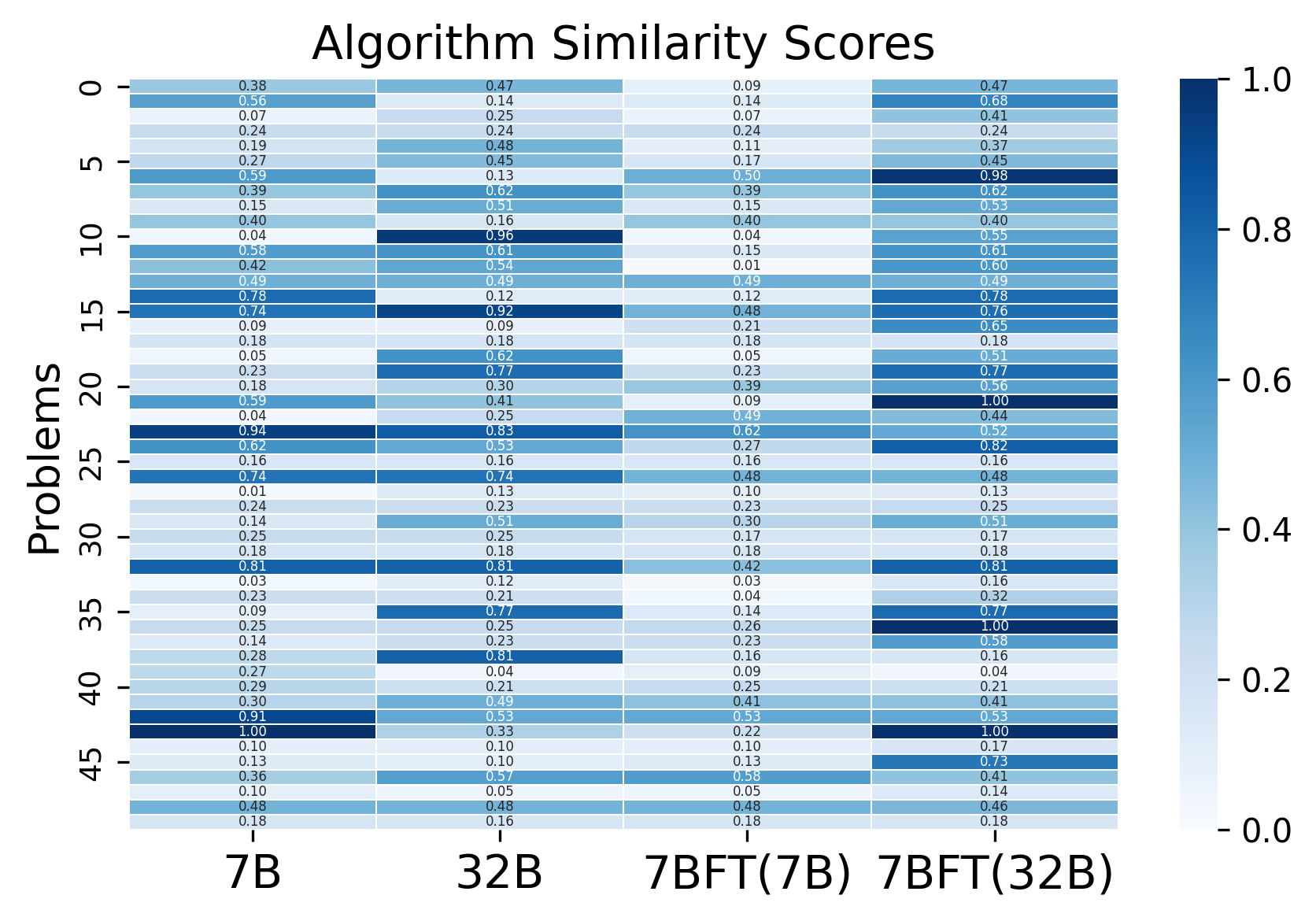}
\caption{
\textcolor{blue}{Cosine similarity between algorithm descriptions generated by base and fine-tuned models and ground-truth algorithm tags for 50 xCodeEval problems.} Darker cells indicate stronger alignment.
}
\label{fig:algorithm-quality}
\end{figure}

\textcolor{blue}{Our analysis further shows that self-collected trajectories from a 7B model often mislabel tasks, such as overpredicting dynamic programming, and exhibit limited algorithmic diversity. As a result, fine-tuning on these traces leads to lower performance (7BFT). In contrast, trajectories generated by 32B align more closely with ground-truth tags and provide more accurate algorithm descriptions. Fine-tuning on these trajectories (7BFT(32B)) significantly improves performance, demonstrating that strong-LLM supervision transfers both correctness and algorithmic diversity to smaller models.}


\begin{figure*}[t]
\centering
\includegraphics[width=\textwidth]{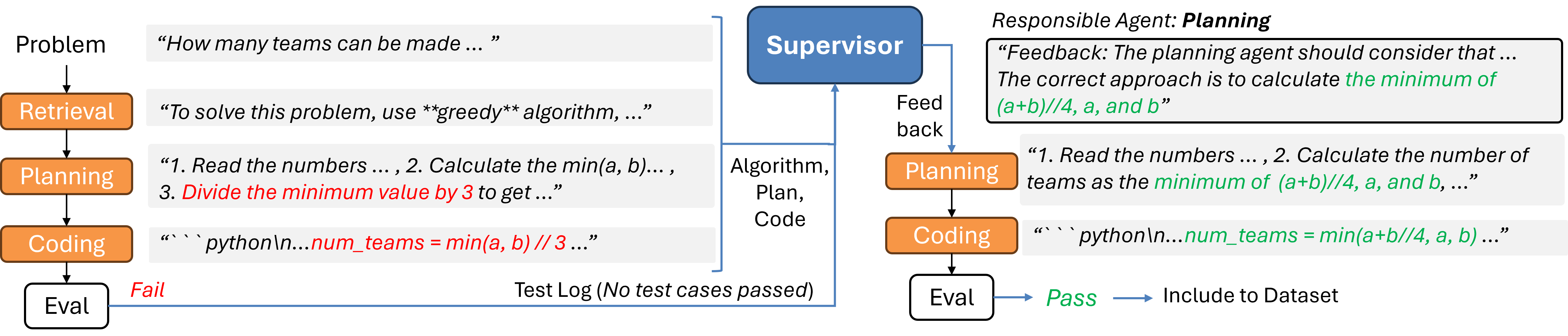}
\caption{Supervisor-aided data collection pipeline. When the final output fails, the supervisor inspects the full trajectory (including algorithm, plan, code, and test result), identifies the responsible agent, and provides targeted feedback to revise its output. If the revised result passes, the updated trajectory is added to the fine-tuning dataset.}
\label{fig:supervisor-pipeline}
\end{figure*}

\subsection{Supervisor-Guided Planning and Coding}
\label{subsec:method2}

Even when fine-tuned on strong-model trajectories, 7B agents struggle to achieve robust multi-stage reasoning. One challenge is the lack of \textit{global awareness}, where agents trained in isolation fail to account for downstream dependencies, an issue previously discussed in Section~\ref{sec:challenge}. Another is the \textit{capacity gap}~\cite{bansal2024smallerweakerbettertraining, xu2025speculativeknowledgedistillation}: while strong LLMs produce coherent and mostly correct outputs, small models tend to mimic surface forms without learning the underlying reasoning. These limitations underscore the need for supervision that supports both global coordination and deeper abstraction.

To address these issues, we propose a \textit{supervisor-guided refinement pipeline} that supplies global feedback without enlarging the runtime model. As shown in Fig.~\ref{fig:supervisor-pipeline}, a 7B MapCoder first produces a complete retrieval→planning→coding trajectory.  If the resulting program fails its unit tests, the entire trajectory is forwarded to a high-capacity \textit{supervisor} LLM (DeepSeek-V3). \textcolor{blue}{The supervisor analyzes the trace, identifies the agent primarily responsible for the failure, and issues concise, role-specific feedback. Only the selected agent then regenerates its output.} The revised trajectory is re-tested; once all tests pass, the final plan–code pair is added to the fine-tuning corpus. 

Supervisor-guided refinement is applied selectively to trajectories where the strong LLM succeeds but the 7B model fails, minimizing generation overhead. The supervisor operates \emph{only} during data generation, keeping inference lightweight. We store only the corrected agent input–output pairs, omitting the feedback itself, so the 7B model learns solely from information available at runtime. Crucially, every example in this corpus is produced and execution-validated by the 7B model, eliminating concerns about a capacity gap. Iterating over thousands of problems yields a dataset that better aligns planners and coders with end-to-end success. Table~\ref{tab:supervisor-ablation} confirms that this strategy delivers a substantial accuracy boost.

\subsection{Multi-Agent LoRA Fine-Tuning}
\label{subsec:method3}

\begin{table}[t]
\centering
\resizebox{\linewidth}{!}{%
\begin{tabular}{l|c|c|c}
\toprule
\textbf{FT Method} & \makecell{\textbf{Trainable} \\ \textbf{Parameters}} & \makecell{\textbf{Required Memory} \\ \textbf{for Training (GB)}} & \makecell{\textbf{Accuracy} \\ (\%)} \\
\midrule
FFT  & 7615.62M  & 45.69 & 18.87 \\
LoRA & 20.19M & 15.35  & 22.64 \\
\bottomrule
\end{tabular}
}
\caption{The number of trainable parameters, required memory for training, and xCodeEval accuracy of full fine-tuning (FFT) and LoRA methods with MapCoder-Lite 7B. Accuracy is measured only up to the coding stage, without debugging.}
\label{tab:lora-vs-fft}
\end{table}

\begin{figure}[t]
\centering
\includegraphics[width=\linewidth]{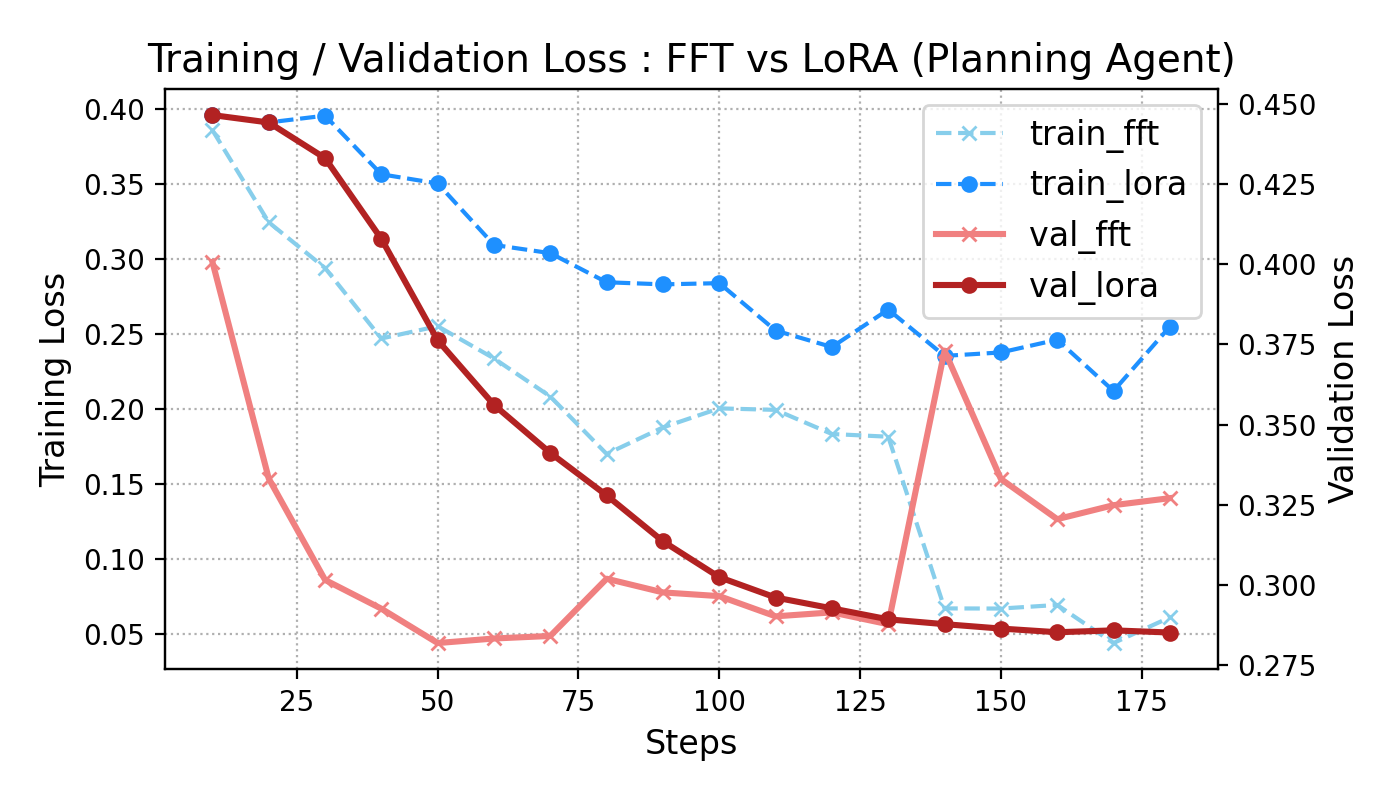}
\caption{
Training and validation loss for the Planning agent using full fine-tuning (FFT) and LoRA. LoRA shows higher training loss but maintains lower and more stable validation loss, suggesting better generalization.
}
\label{fig:train-loss}
\end{figure}


To enable role-specific specialization, we adopt an \textit{agent-wise LoRA} strategy: all four agents share a frozen 7B backbone, each with independent rank-32 LoRA adapters fine-tuned on role-specific data. Unlike prior approaches that fine-tune the entire model across tasks~\cite{shen-etal-2024-small, zeng2023agenttuningenablinggeneralizedagent}, our method isolates agent behavior at the adapter level while preserving a unified backbone.

\begin{table*}[t]
\centering
\resizebox{\textwidth}{!}{%
\begin{tabular}{c|ccc|ccc|ccc}
\toprule
\multirow{2}{*}{\textbf{Method}} & \multicolumn{3}{c|}{\textbf{xCodeEval (106)}} & \multicolumn{3}{c|}{\textbf{APPS (150)}} & \multicolumn{3}{c}{\textbf{CodeContests (165)}} \\
\cmidrule(lr){2-4} \cmidrule(lr){5-7} \cmidrule(l){8-10}
 & Accuracy$\uparrow$ & Pass Count$\uparrow$ & Format Fails$\downarrow$ & Accuracy$\uparrow$ & Pass Count$\uparrow$ & Format Fails$\downarrow$ & Accuracy$\uparrow$ & Pass Count$\uparrow$ & Format Fails$\downarrow$ \\
\midrule
Direct        & 18.87 & 20 & -- & 4.00 & 6  & -- & 1.21  & 2  & -- \\
CoT           & 5.66  & 6  & -- & 4.67 & 7  & -- & 3.64  & 6  & -- \\
Self-planning & 9.43  & 10 & -- & 1.33 & 2  & -- & 3.03  & 5  & -- \\
Analogical    & 12.26 & 13 & -- & 3.33 & 5  & -- & 1.82  & 3  & -- \\
MapCoder      & 13.21 & 14 & 29 & 6.00 & 9  & 30 & 6.06  & 10 & 59 \\
MapCoder-Lite   & \textbf{28.30} & \textbf{29} &  \textbf{0} & \textbf{8.00} & \textbf{12} &  \textbf{0} & \textbf{13.33} & \textbf{22} &  \textbf{0} \\
\bottomrule
\end{tabular}
}
\caption{Performance comparison of different methods using Qwen7B on xCodeEval, APPS, and CodeContests with Pass@1 accuracy (\%). The numbers in parentheses indicate the number of problems in each benchmark.}
\label{tab:main-results}
\end{table*}

We empirically show that LoRA serves effectively as a modular control layer in multi-agent code generation, delivering both accuracy and efficiency gains (Table~\ref{tab:lora-vs-fft}). We attribute this to LoRA’s implicit regularization~\cite{hu2021lora, houlsby2019parameterefficient}, which promotes task-specific behavior without overwriting pretrained knowledge. In multi-agent settings with complex, structured I/O, full fine-tuning often overfits to superficial patterns, weakening general reasoning. As shown in Fig.~\ref{fig:train-loss}, full fine-tuning achieves lower training loss but increasing validation loss, whereas LoRA maintains lower and more stable validation loss, indicating better generalization and adaptability across roles.


\section{Experimental Settings}


\textbf{Models.}
All agent-wise fine-tuning experiments use Qwen2.5-7B-Instruct as the base model. For comparison, we include larger variants—Qwen2.5-14B-Instruct and Qwen2.5-32B-Instruct—which serve as upper bounds for evaluating the effectiveness of our lightweight strategy. To assess whether a code-pretrained backbone improves performance, we additionally test Qwen2.5-Coder-7B-Instruct. We also apply our method to Llama3.1-8B and CodeLlama-7B to demonstrate generalizability beyond the Qwen family.

\textbf{Benchmarks.}
We evaluate models on three competitive programming benchmarks: xCodeEval~\cite{khan-etal-2024-xcodeeval}, APPS~\cite{hendrycks2021apps}, and CodeContests~\cite{doi:10.1126/science.abq1158_codecontest}. For xCodeEval, we use the 106 problems from the compact program synthesis split. For CodeContests, we evaluate all 165 official test problems. For APPS, we randomly select 150 test-set problems, following MapCoder~\cite{islam-etal-2024-mapcoder}. Each benchmark provides natural language problem descriptions with sample I/O and is evaluated using hidden unit tests. We additionally report results on HumanEval and MBPP to assess generalization to function-level tasks.


\textbf{Fine-Tuning Setup.}
Each agent is fine-tuned using LoRA adapters with rank-32, applied to the linear layers for query, key, value, and output projections. We use a learning rate of 2e-5, gradient accumulation of 16, and train for three epochs, with hyperparameters tuned per agent to ensure stability. All training runs are conducted on a single NVIDIA A100 80GB GPU. The cost of data curation prior to fine-tuning is summarized in Table~\ref{tab:data} (Section~\ref{sec:data}).

\textbf{Evaluation Setting.}
We report pass@1 accuracy using greedy decoding to ensure reproducibility and align with baseline evaluations.


\section{Results}

\subsection{Effectiveness of Agent-wise Fine-Tuning}

Table~\ref{tab:main-results} compares the performance of Qwen2.5-7B under various prompting strategies—direct prompting, CoT\cite{wei2023chainofthought}, self-planning\cite{jiang2024selfplanning}, analogical~\cite{yasunaga2024analogical}—and multi-agent setups. Without fine-tuning, MapCoder yields only marginal improvements and even underperforms direct prompting on xCodeEval (13.21\% vs. 18.87\%), underscoring the difficulty of coordinating small models in multi-agent pipelines due to format errors and poor inter-agent coordination.

In contrast, MapCoder-Lite with agent-wise fine-tuning achieves substantial improvements across all benchmarks. On xCodeEval, accuracy more than doubles to 28.30\%, outperforming all prompting baselines. Similar gains are observed on APPS (6.00\% → 8.00\%) and CodeContests (6.06\% → 13.33\%). Notably, fine-tuning eliminates all format failures, indicating that even small LMs can produce structurally consistent outputs when role-specialized. These results highlight the critical role of agent-wise fine-tuning in realizing the full potential of small LMs in multi-agent frameworks, with qualitative examples in Appendix~\ref{sec:improvements} further illustrating improvements across all roles.

\subsection{Comparison with Larger Models}  
Table~\ref{tab:scale-results} compares MapCoder's performance across backbone sizes from 7B to 32B (Qwen[7–32]B), alongside MapCoder-Lite—our fine-tuned 7B variant (Qwen7B-FT). Qwen7B-FT delivers a substantial accuracy boost over the untuned Qwen7B, matching the performance of Qwen14B and coming within six points of Qwen32B. This is particularly notable given that Qwen7B-FT requires only one-quarter of the GPU memory of Qwen32B, making it viable for deployment on memory-constrained edge devices~\cite{Karumbunathan2022JetsonAGXOrin}. Moreover, the reduced model size yields a proportional speedup in token generation—LLM decoding being memory-bound—achieving a 4× reduction in time-per-output-token (TPOT), as measured using vLLM~\cite{kwon2023efficientmemorymanagementlarge}. These results highlight that targeted fine-tuning enables small models to achieve competitive performance with dramatically lower cost.

\begin{table}[t]
\centering
\resizebox{\linewidth}{!}{%
\begin{tabular}{l|ccc|cc}
\toprule
\textbf{Model} & \textbf{xCodeEval} & \textbf{APPS} & \textbf{CodeContests} & \textbf{Memory} & \textbf{TPOT} \\
\midrule
Qwen7B           & 13.21 & 6.00  & 6.06  & 15.26 & 12.29 \\
\textbf{Qwen7B-FT} & \textbf{28.30} & \textbf{8.00}  & \textbf{13.33} & \textbf{15.64} & \textbf{12.29} \\
Qwen14B          & 28.30 & 10.00 & 15.76 & 29.58 & 23.36 \\
Qwen32B          & 33.02 & 13.33 & 18.18 & 65.56 & 45.06 \\
\bottomrule
\end{tabular}
}
\caption{Pass@1 accuracy (\%) of MapCoder using Qwen models of different scales on xCodeEval, APPS, and CodeContests. 
Memory usage (GB) and Time Per Output Tokens (ms).}
\label{tab:scale-results}
\end{table}


\subsection{Ablation Study}
\textbf{Contribution of Individual Agent Fine-Tuning.}  
Table~\ref{tab:ablation-agent} presents an ablation study of agent-wise fine-tuning on xCodeEval. Tuning the retrieval agent alone raises accuracy from 13.21\% to 18.87\% and cuts format failures from 29 to 12. Adding the planning agent further improves accuracy to 26.42\% and eliminates all format errors, boosting both non-debug and debug-assisted completions. Fine-tuning the coding agent (without debugging) increases non-debug passes to 24 but reduces debug-assisted ones, as stronger initial code often bypasses the debugger. In contrast, tuning the debugging agent (without coding) improves recovery from weak code, achieving 27.36\% accuracy and 12 debug-assisted completions. Full fine-tuning of all agents yields the best performance (28.30\%), confirming that each agent contributes uniquely and coordinated tuning is critical for optimal results.

\begin{table}[t]
\centering
\resizebox{\linewidth}{!}{%
\begin{tabular}{cccc|cccc}
\toprule
\multicolumn{4}{c|}{\textbf{Agent}} & \textbf{xCodeEval} & \textbf{Pass} & \textbf{Pass} & \textbf{Format} \\
\textbf{R} & \textbf{P} & \textbf{C} & \textbf{D} & \textbf{Accuracy (\%)} & \textbf{w/o Debug$\uparrow$} & \textbf{w/ Debug$\uparrow$} & \textbf{Fail$\downarrow$} \\
\midrule
- & - & - & - & 13.21 & 11 & 3  & 29 \\
\ding{51} & - & - & - & 18.87 & 17 & 3  & 12 \\
\ding{51} & \ding{51} & - & - & 26.42 & 18 & 10 & 0  \\
\ding{51} & \ding{51} & \ding{51} & - & 24.53 & 24 & 2  & 0  \\
\ding{51} & \ding{51} & - & \ding{51} & 27.36 & 17 & 12 & 0  \\
\ding{51} & \ding{51} & \ding{51} & \ding{51} & \textbf{28.30} & \textbf{24} & \textbf{6} & \textbf{0} \\
\bottomrule
\end{tabular}
}
\caption{Ablation study of per-agent fine-tuning contributions in MapCoder on xCodeEval. R, P, C, and D denote Retrieval, Planning, Coding, and Debugging agents, respectively.}
\label{tab:ablation-agent}
\end{table}

\begin{table}[t]
\centering
\small
\begin{tabular}{ccc|c}
\toprule
\textbf{Retrieval} & \textbf{Planning} & \textbf{Coding} & \textbf{xCodeEval (\%)} \\
\midrule
32 & 32 & 32 & \textbf{22.64} \\
8  & 32 & 32 & 18.87 \\
32 & 8  & 32 & 20.75 \\
32 & 32 & 8  & 21.70 \\
8  & 8  & 8  & 21.70 \\
64 & 32 & 32 & 16.98 \\
32 & 64 & 32 & 20.75 \\
32 & 32 & 64 & 20.75 \\
64 & 64 & 64 & 16.98 \\
\bottomrule
\end{tabular}
\caption{\textcolor{blue}{Effect of LoRA rank on multi-agent performance. The three leftmost columns indicate the LoRA ranks applied to the retrieval, planning, and coding agents, respectively.}}
\label{tab:lora_rank_ablation}
\end{table}

\begin{table*}[t]
\centering
\resizebox{\textwidth}{!}{%
\begin{tabular}{c|cccc|cccc}
\toprule
\multirow{2}{*}{\textbf{Method}} & \multicolumn{4}{c|}{\textbf{HumanEval (164 problems)}} & \multicolumn{4}{c}{\textbf{MBPP (397 problems)}} \\
\cline{2-9}
 & Accuracy (\%) & Pass w/o Debug$\uparrow$ & Pass w/ Debug$\uparrow$ & Format Fails$\downarrow$ & Accuracy (\%) & Pass w/o Debug$\uparrow$ & Pass w/ Debug$\uparrow$ & Format Fails$\downarrow$ \\
\midrule
MapCoder       & 70.73 & 105 & 11 & 14 & 67.51 & 244 & 24 & 68 \\
MapCoder-Lite  & 82.93 & 120 & 16 & 1  & 84.63 & 305 & 31 & 0 \\
\bottomrule
\end{tabular}
}
\caption{Evaluation of MapCoder and MapCoder-Lite on HumanEval and MBPP benchmarks.}
\label{tab:functional-eval}
\end{table*}

\textcolor{blue}{
\textbf{Impact of LoRA Rank.}
We study the effect of LoRA rank on multi-agent performance to justify the configuration used in this work. The rank-32 setting was selected based on empirical evaluation rather than by convention. As shown in Table~\ref{tab:lora_rank_ablation}, rank-32 achieves the highest xCodeEval accuracy (22.64\%) among all tested configurations. Lower-rank settings (e.g., $(8,8,8)$) underperform despite being parameter-efficient, while higher-rank configurations (e.g., 64) do not yield further gains and can degrade performance. These results suggest that multi-agent performance depends on interactions between agent roles rather than parameter count alone. Overall, rank-32 offers the best balance between accuracy and efficiency, introducing only about 3\% additional trainable parameters relative to the backbone.
}

\textbf{Backbone Selection.}
Choosing the right backbone is critical for multi-agent code generation, where each agent performs a distinct, reasoning-intensive role. Among SLM candidates, we compare a general-purpose model (Qwen2.5-7B-Instruct~\cite{qwen2025qwen25technicalreport}) and a code-specialized variant (Qwen2.5-Coder-7B-Instruct~\cite{hui2024qwen25coder}) for use in MapCoder. Despite similar sizes, the general-purpose model achieves higher accuracy and format adherence. Agent-wise ablations further show that replacing even one agent with its coder counterpart degrades performance, suggesting that the coder model lacks the contextual reasoning and adaptability needed for role-specific tasks. These results support our choice of the general-purpose model as the unified backbone for MapCoder-Lite (see Appendix~\ref{sec:coder}).

\subsection{\textcolor{blue}{Generalization and Robustness}}
\textbf{Evaluation on Simpler Coding Tasks.}  
To evaluate generalization beyond competition-level programming, we compare MapCoder and MapCoder-Lite on two function-level benchmarks: HumanEval and MBPP. As shown in Table~\ref{tab:functional-eval}, MapCoder-Lite outperforms the baseline by significantly reducing format-related failures and achieving higher end-to-end accuracy. It improves both coding (increased passes without debugging) and debugging (even greater gains in passes with debugging), indicating enhanced reliability. These results suggest that MapCoder-Lite not only preserves generalization capability but also improves robustness on structurally simpler tasks.

\textbf{Model Compatibility.}
\textcolor{blue}{Our method is architecture-agnostic and applicable to LoRA-supported backbones. Beyond the Qwen2.5 family, MapCoder-Lite improves accuracy and reduces format failures on Llama3.1-8B and CodeLlama-7B across xCodeEval and HumanEval (Appendix~\ref{sec:compatibility}). We also evaluate MapCoder-Lite on the recent reasoning-oriented model Qwen3-4B; despite strong single-agent prompting performance, stable multi-agent behavior is only achieved after agent-wise fine-tuning, with detailed results reported in Appendix~\ref{sec:qwen3}.}

\subsection{\textcolor{blue}{Data Curation Cost and Statistics}}
\label{sec:data}
\textcolor{blue}{Table~\ref{tab:data} summarizes the dataset size, resource usage, and data generation time for each agent. In particular, we curate approximately 2.3k execution-verified trajectories for the retrieval agent, 1.1k each for the planning and coding agents, and 4.2k for the debugging agent. We used 2×A100 GPUs with vLLM serving to run Qwen2.5-32B and Qwen2.5-7B models, and partially relied on the DeepSeek API as supervisor and debugging agent, incurring \$20–30. Data generation is performed only once, and the resulting datasets are reusable across models and experiments. The strong supervisor model is invoked only in failure cases to provide brief corrective feedback, rather than full trajectory generation, keeping strong-model involvement infrequent and tightly bounded to training-time supervision. Retrieval and debugging took longer due to longer outputs; debugging in particular depends on planning and coding outputs, requiring upstream completion. While data was generated per agent during development, the pipeline can be streamlined into a single pass for faster future use.}

\begin{table}[t]
\centering
\resizebox{\linewidth}{!}{%
\begin{tabular}{lccc}
\hline
 & \textbf{Retrieval} & \textbf{Planning, Coding} & \textbf{Debugging} \\
\hline
Dataset size & 2.3k & 1.1k (each) & 4.2k \\
Time & 2--3 days & 16 hours & 4--5 days \\
GPUs & 2 $\times$ A100 & 2 $\times$ A100 & 2 $\times$ A100 \\
\hline
\end{tabular}
}
\caption{\textcolor{blue}{Dataset size, data generation time, and computational resource usage for data curation for each agent.}}
\label{tab:data}
\end{table}

\section{Conclusion}


We introduced \emph{MapCoder-Lite}, a multi-agent coding framework built on a 7B model with lightweight LoRA adapters. \textcolor{blue}{Through trajectory distillation, supervisor-guided refinement, and agent-wise specialization, it achieves competitive performance at substantially lower cost. Our results demonstrate that small language models, when carefully fine-tuned, can match the reliability and accuracy of much larger systems in multi-agent code generation.}

\section*{Acknowledgments}

This work was supported by the National Research Foundation of Korea (NRF) grant funded by the Korea government (MSIT) (No. RS-2025-00561961). This work was also supported by the Institute of Information \& Communications Technology Planning \& Evaluation (IITP) grant funded by the Korea government (MSIT) (No. RS-2025-02214497, Development of Low-Level Optimization Program API Technology for AI Semiconductors).

\section*{Limitations}

Our framework relies on distilled trajectories from strong models such as DeepSeek-V3 and Qwen2.5-32B, yet the fine-tuned 7B model does not fully replicate their performance. Rather than indicating a hard performance ceiling, this gap highlights opportunities to further enhance small models, potentially through architectural extensions or improved training objectives. In addition, the multi-agent structure introduces a broad design space for fine-tuning, as each agent has a distinct role and optimization objective. While we adopt a uniform tuning recipe across agents in this work, more adaptive tuning could further improve performance.


\bibliography{refs}

\clearpage

\appendix

\section{Details of Supervisor}

\begin{figure*}[t]
\centering
\includegraphics[width=\textwidth]{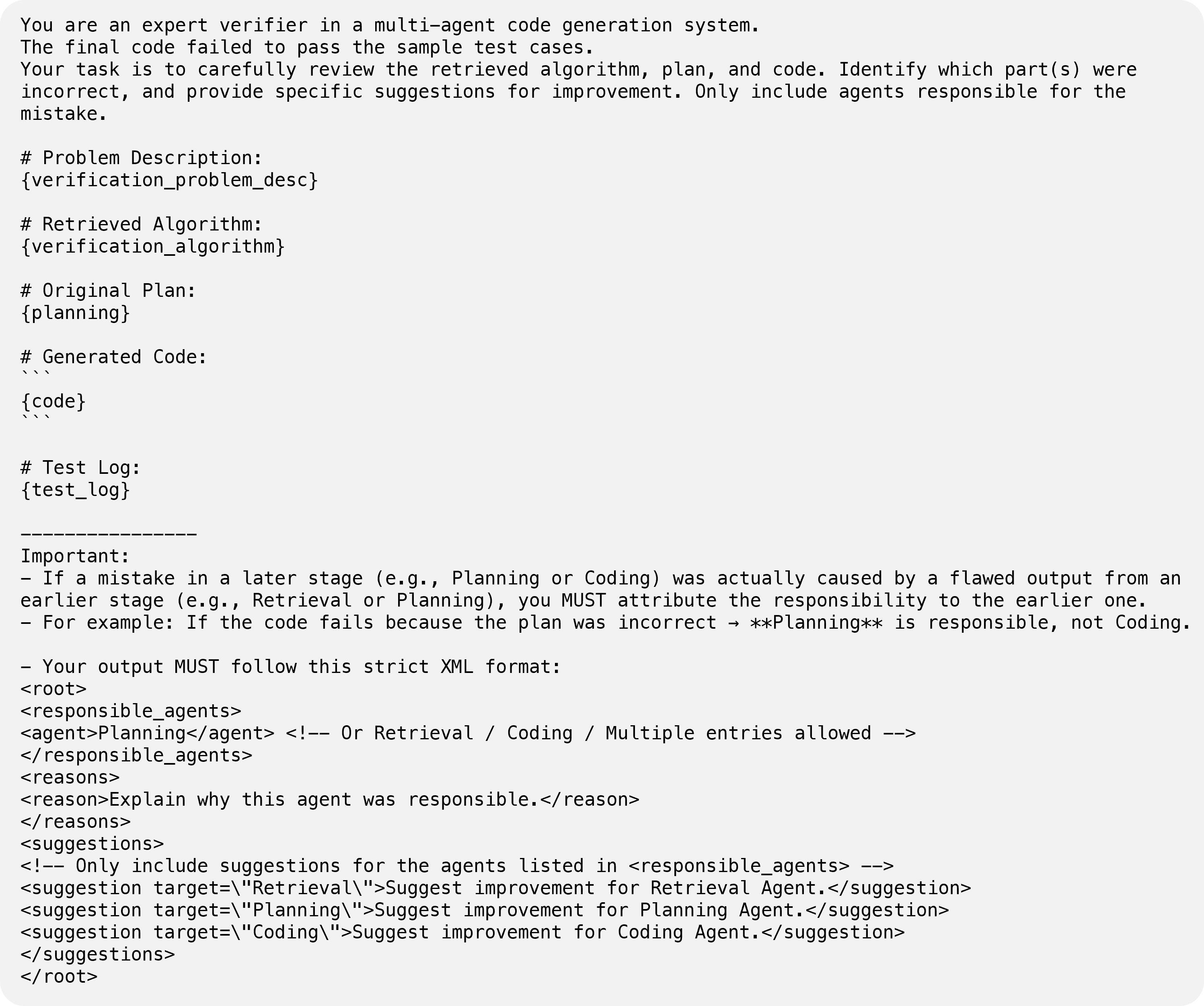}
\caption{Prompt for the Supervisor.}
\label{fig:supervisor_prompt}
\end{figure*}

\begin{figure*}[t]
\centering
\includegraphics[width=\textwidth]{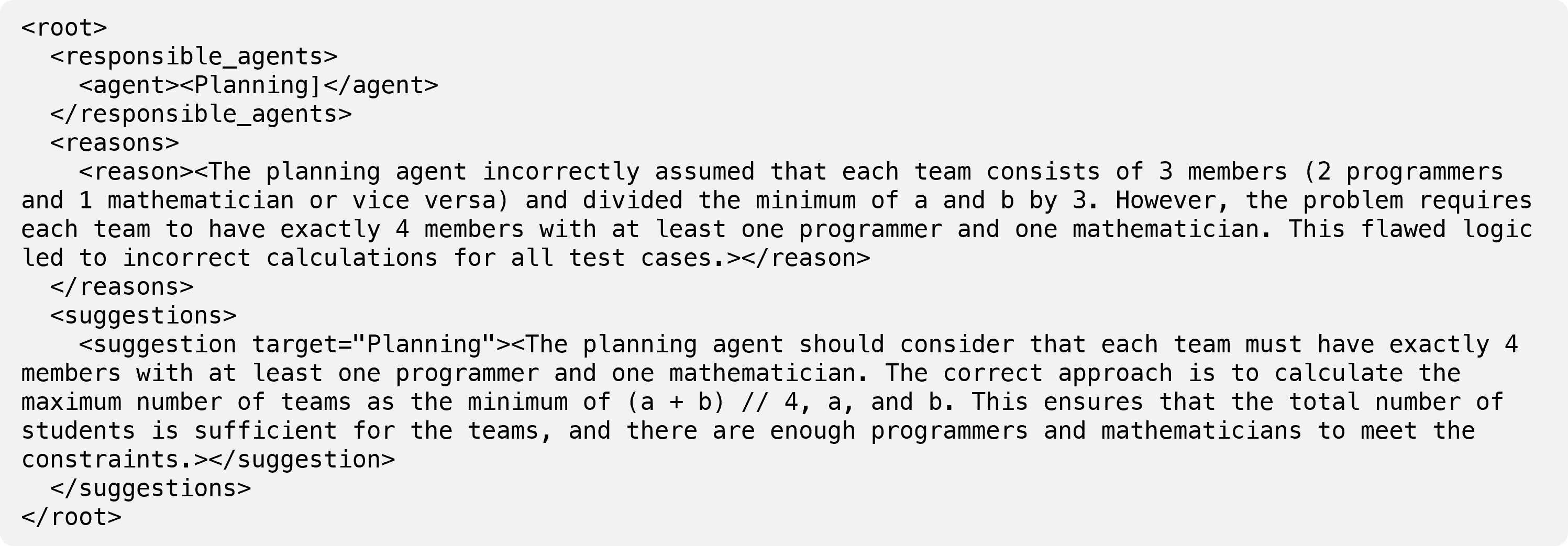}
\caption{Example response from the supervisor: identifies the flawed part of the plan and generates feedback.}
\label{fig:supervisor_response}
\end{figure*}

Fig.~\ref{fig:supervisor_prompt} and Fig.~\ref{fig:supervisor_response} illustrate the operation of the supervisor in our data curation pipeline. The supervisor receives the full execution trajectory of a problem, including the problem description, intermediate agent outputs (e.g., algorithm explanation, step-by-step plan, and code), and the result of test case execution.

As shown in Fig.~\ref{fig:supervisor_prompt}, the input prompt instructs the supervisor to analyze the trajectory, determine which agent is responsible for the failure, and provide natural language feedback targeting that agent. The goal is to isolate the error at the correct stage (retrieval, planning, or coding), and produce actionable guidance that enables re-generation only the faulty component.

Fig.~\ref{fig:supervisor_response} shows a representative response from the supervisor. It correctly attributes the failure to the planning agent, explains why the plan is insufficient, and suggests how it should be modified. This output is then used to re-invoke the corresponding agent with additional guidance, creating high-quality training data without human annotation.

\section{Backbone Model Comparison: General-Purpose vs. Coder}
\label{sec:coder}
To investigate the impact of backbone model selection in multi-agent code generation, we conducted a series of controlled experiments comparing general-purpose and code-specialized models. Specifically, we tested Qwen2.5-7B-Instruct (general-purpose) and Qwen2.5-Coder-7B-Instruct (code-specialized) under various configurations of the MapCoder and MapCoder-Lite pipelines. The results show that general-purpose models are more robust across all agent roles and better suited for both zero-shot and fine-tuned multi-agent setups.

\textbf{Benchmark Performance (Zero-shot).} Table~\ref{tab:model-benchmark} shows the performance of MapCoder using each backbone model. The general-purpose variant consistently achieves higher accuracy and fewer format failures, highlighting its superiority in multi-agent coordination and instruction-following tasks.


\begin{table}[t]
\centering
\resizebox{\linewidth}{!}{%
\begin{tabular}{l|l|cc}
\toprule
\textbf{Benchmark} & \textbf{Metric} & \textbf{7B} & \textbf{7BCoder} \\
\midrule
\multirow{2}{*}{xCodeEval (106)}     & Accuracy (\%)$\uparrow$     & 13.21 & 8.49 \\
                                     & Format Fails$\downarrow$    & 29    & 44   \\
\midrule
\multirow{2}{*}{APPS (150)}          & Accuracy (\%)$\uparrow$     & 6.00  & 2.49 \\
                                     & Format Fails$\downarrow$    & 30    & 66   \\
\midrule
\multirow{2}{*}{CodeContests (165)}  & Accuracy (\%)$\uparrow$     & 6.06  & 5.45 \\
                                     & Format Fails$\downarrow$    & 59    & 77   \\
\bottomrule
\end{tabular}
}
\caption{Comparison of Qwen2.5-7B-Instruct and Qwen2.5-Coder-7B-Instruct across benchmarks.}
\label{tab:model-benchmark}
\end{table}

\textbf{Agent-wise Ablation (Zero-shot).} As shown in Table~\ref{tab:coder-ablation-zero}, replacing any agent in the general-purpose pipeline with its coder counterpart leads to a consistent accuracy drop, most notably in the planning and coding agents. Even for the coding role, where the coder model is expected to excel, performance decreases—highlighting the importance of upstream integration and general reasoning.

\begin{table}[t]
\centering
\resizebox{\linewidth}{!}{%
\begin{tabular}{cccc|c}
\toprule
\textbf{Retrieval} & \textbf{Planning} & \textbf{Coding} & \textbf{Debugging} & \textbf{xCodeEval (\%)} \\
\midrule
7B & 7B & 7B & 7B & 13.21 \\
7BCoder & 7B & 7B & 7B & 12.26 \\
7B & 7BCoder & 7B & 7B & 10.38 \\
7B & 7B & 7BCoder & 7B & 10.38 \\
7B & 7B & 7B & 7BCoder & 12.26 \\
\bottomrule
\end{tabular}
}
\caption{
Ablation of 7B vs. 7B Coder in retrieval, planning, coding, and debugging agents on the xCodeEval benchmark.}
\label{tab:coder-ablation-zero}
\end{table}

\textbf{Training Loss after Fine-tuning.} We fine-tuned both models using identical LoRA settings across all agents. As summarized in Table~\ref{tab:training-loss}, the coder model converges to higher final loss values, suggesting a poorer fit to role-specialized training data. This pattern is consistent across all agents.

\begin{table}[t]
\centering
\resizebox{\linewidth}{!}{%
\begin{tabular}{lcccc}
\toprule
\textbf{Model} & \textbf{Retrieval} & \textbf{Planning} & \textbf{Coding} & \textbf{Debugging} \\
\midrule
Qwen2.5-7B-Instruct       & 0.34 & 0.25 & 0.08 & 0.33 \\
Qwen2.5-Coder-7B-Instruct & 0.36 & 0.35 & 0.13 & 0.34 \\
\bottomrule
\end{tabular}
}
\caption{
Final training loss after LoRA fine-tuning for each agent using general-purpose vs.\ coder-specific models.}
\label{tab:training-loss}
\end{table}

\textbf{Ablation after LoRA Fine-tuning.} Table~\ref{tab:coder-ablation-ft} presents accuracy when substituting each LoRA-finetuned agent with its coder-based variant. Accuracy declines in all settings, and the fully coder-based pipeline achieves only 18.87\%, compared to 28.30\% for the general-purpose variant. These results indicate that even after fine-tuning, general-purpose models better support the multi-agent workflow.

\begin{table}[t]
\centering
\resizebox{\linewidth}{!}{%
\begin{tabular}{cccc|c}
\toprule
\textbf{Retrieval} & \textbf{Planning} & \textbf{Coding} & \textbf{Debugging} & \textbf{xCodeEval (\%)} \\
\midrule
7BFT & 7BFT & 7BFT & 7BFT & 28.30 \\
7BCoderFT & 7BFT & 7BFT & 7BFT & 21.70 \\
7BFT & 7BCoderFT & 7BFT & 7BFT & 23.58 \\
7BFT & 7BFT & 7BCoderFT & 7BFT & 24.53 \\
7BFT & 7BFT & 7BFT & 7BCoderFT & 23.58 \\
7BCoderFT & 7BCoderFT & 7BCoderFT & 7BCoderFT & 18.87 \\
\bottomrule
\end{tabular}
}
\caption{
xCodeEval accuracy when substituting each agent with a LoRA-finetuned coder-specific model (7BCoderFT), compared against the general-purpose baseline (7BFT).
}
\label{tab:coder-ablation-ft}
\end{table}

\section{Evaluation on Recent Reasoning Models (Qwen3)}
\label{sec:qwen3}

\textcolor{blue}{
We evaluate MapCoder-Lite on Qwen3-4B, a recently released small language model that significantly improves single-agent reasoning and outperforms Qwen2.5-7B under direct prompting. As shown in Table~\ref{tab:qwen3_comparison}, Qwen3-4B achieves strong direct-prompting performance (22.64\% without thinking and 26.42\% with thinking). However, this improvement does not directly translate into reliable multi-agent behavior. When deployed in the MapCoder pipeline without fine-tuning, Qwen3-4B attains only 3.77\% accuracy, exhibiting the same instability observed with earlier backbones.}

\textcolor{blue}{
Applying MapCoder-Lite mitigates this issue. After fine-tuning the retrieval, planning, and coding agents, Qwen3-4B achieves 30.19\% accuracy, surpassing the fine-tuned Qwen2.5-7B despite using fewer parameters. These results indicate that recent advances in single-agent reasoning alone are insufficient for stable multi-agent workflows, and that targeted agent-wise specialization remains essential. Moreover, the larger gains observed with Qwen3-4B suggest that MapCoder-Lite scales positively with backbone capability, effectively activating role-specific behaviors that are not induced by prompting alone.
}

\begin{table}[t]
\centering
\small
\resizebox{\linewidth}{!}{%
\begin{tabular}{lcc}
\toprule
\textbf{Method} & \textbf{Qwen2.5-7B-Instruct} & \textbf{Qwen3-4B} \\
\midrule
Direct (non-thinking) & 18.87 & 22.64 \\
Direct (thinking) & -- & 26.42 \\
MapCoder (RPC) & 11.32 & 3.77 \\
MapCoder-Lite (RPC) & 22.64 & \textbf{30.19} \\
\bottomrule
\end{tabular}
}
\caption{xCodeEval performance of Qwen2.5-7B and Qwen3-4B under direct prompting and MapCoder pipelines. RPC denotes the retrieval--planning--coding agents.}
\label{tab:qwen3_comparison}
\end{table}


\section{Generalization Across Model Families}
\label{sec:compatibility}
While our main experiments used the Qwen series due to their strong coding performance on HumanEval and MBPP during development, our method remains architecture-agnostic. Since MapCoder-Lite applies LoRA through the PEFT library, it can be seamlessly used with any compatible model.

To test this generalizability, we evaluated both MapCoder and MapCoder-Lite using Llama3.1-8B-Instruct and CodeLlama-7B-Instruct-hf as backbones. As shown in Table~\ref{tab:cross-model-performance}, both models performed poorly without fine-tuning on competitive programming tasks. For example, Llama3.1-8B scored 0.00\% with 94 format failures on xCodeEval. After applying MapCoder-Lite, Llama3.1-8B achieved 16.04\% accuracy with zero format failures, demonstrating the adaptability of our pipeline. In contrast, CodeLlama-7B showed limited improvement on xCodeEval but responded well on function-level benchmarks such as HumanEval.

\begin{table}[t]
\centering
\resizebox{\linewidth}{!}{%
\begin{tabular}{l|c|c|c}
\toprule
\multirow{2}{*}{\textbf{Model}} & \multirow{2}{*}{\textbf{Benchmark}} 
 & {\textbf{MapCoder}} 
 & {\textbf{MapCoder-Lite}} \\
\cline{3-4}
 & & Accuracy (\%) & Accuracy (\%) \\
\midrule
Llama3.1-8B-Instruct  & xCodeEval & 0.00 & 16.04 \\
CodeLlama-7B-Instruct & xCodeEval & 0.94 & 2.83  \\
CodeLlama-7B-Instruct & HumanEval & 10.98 & 45.12 \\
\bottomrule
\end{tabular}
}
\caption{Comparison of MapCoder and MapCoder-Lite on Llama3.1-8B and CodeLlama-7B models on xCodeEval and HumanEval benchmarks.}
\label{tab:cross-model-performance}
\end{table}

Additionally, we compare MapCoder-Lite with various prompting strategies using the Llama3.1-8B backbone. As summarized in Table~\ref{tab:prompting-comparison}, MapCoder underperforms direct prompting baselines, whereas MapCoder-Lite substantially improves accuracy and surpasses all prompting methods. These results reinforce that multi-agent pipelines require targeted fine-tuning to be effective: without it, they underperform simpler approaches, while with it, they become competitive and scalable.


\begin{table}[t]
\centering
\resizebox{\linewidth}{!}{%
\begin{tabular}{l|c}
\toprule
\textbf{Method} & \textbf{Accuracy (\%)} \\
\midrule
Direct                      & 6.60 \\
CoT                         & 4.72 \\
Self-planning               & 4.72 \\
Analogical                  & 5.66 \\
Multi-agent (MapCoder)      & 0.00 \\
Multi-agent + FT (MapCoder-Lite) & 16.04 \\
\bottomrule
\end{tabular}
}
\caption{Accuracy comparison between MapCoder-Lite and other prompting baselines on xCodeEval using Llama3.1-8B-Instruct.}
\label{tab:prompting-comparison}
\end{table}



\section{Improvements After Fine-Tuning}
\label{sec:improvements}

\begin{figure*}[t]
\centering
\includegraphics[width=\textwidth]{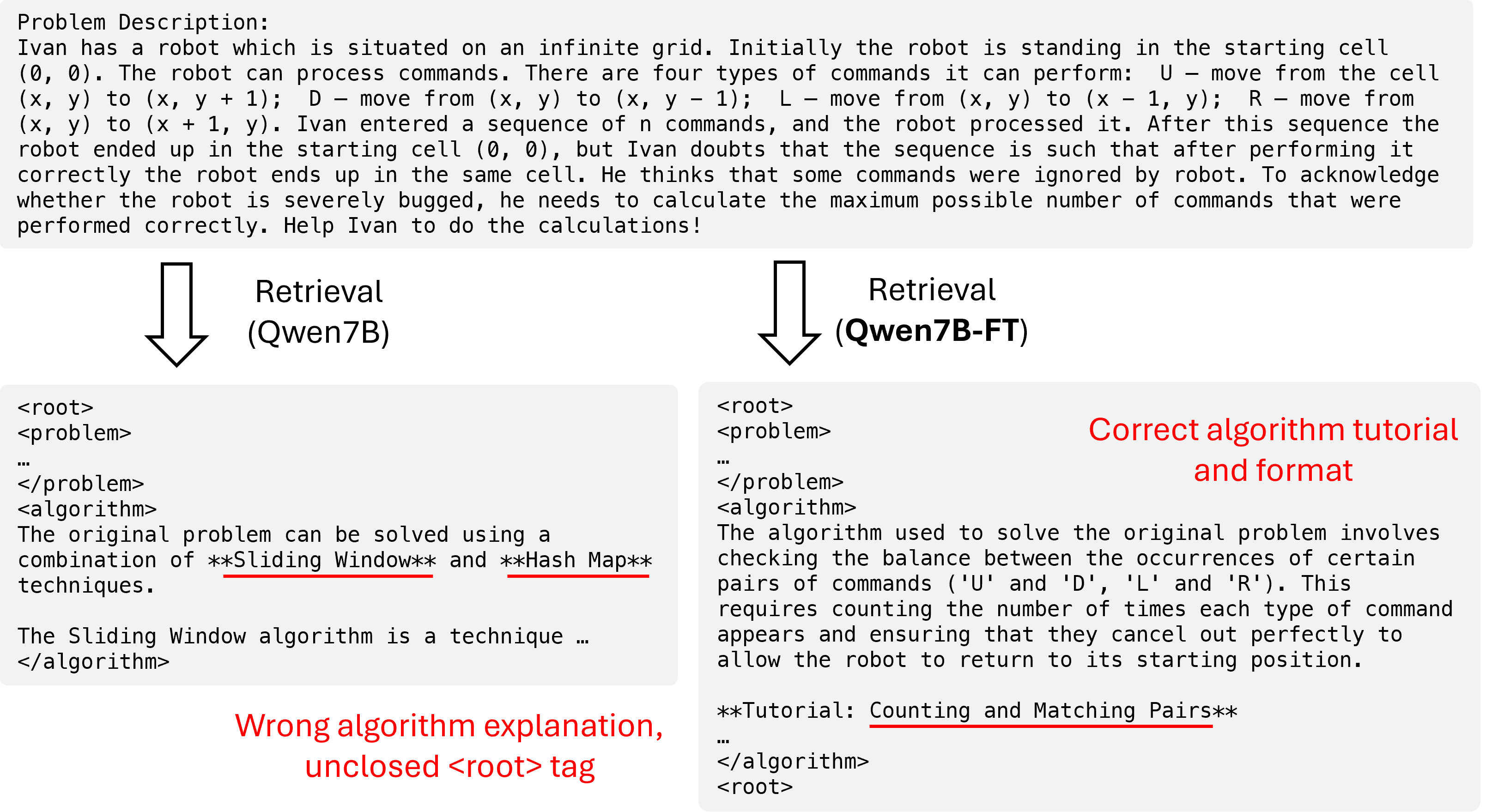}
\caption{Improvement in algorithm tutorial and XML formatting by the retrieval agent after fine-tuning. The pre-trained model fails to identify the correct algorithm and generates an ill-formed XML response with an unclosed \texttt{<root>} tag. After fine-tuning, the retrieval agent provides a valid explanation based on matching command pairs and generates a well-structured XML block.}
\label{fig:retrieval-example}
\end{figure*}

\begin{figure*}[t]
\centering
\includegraphics[width=\textwidth]{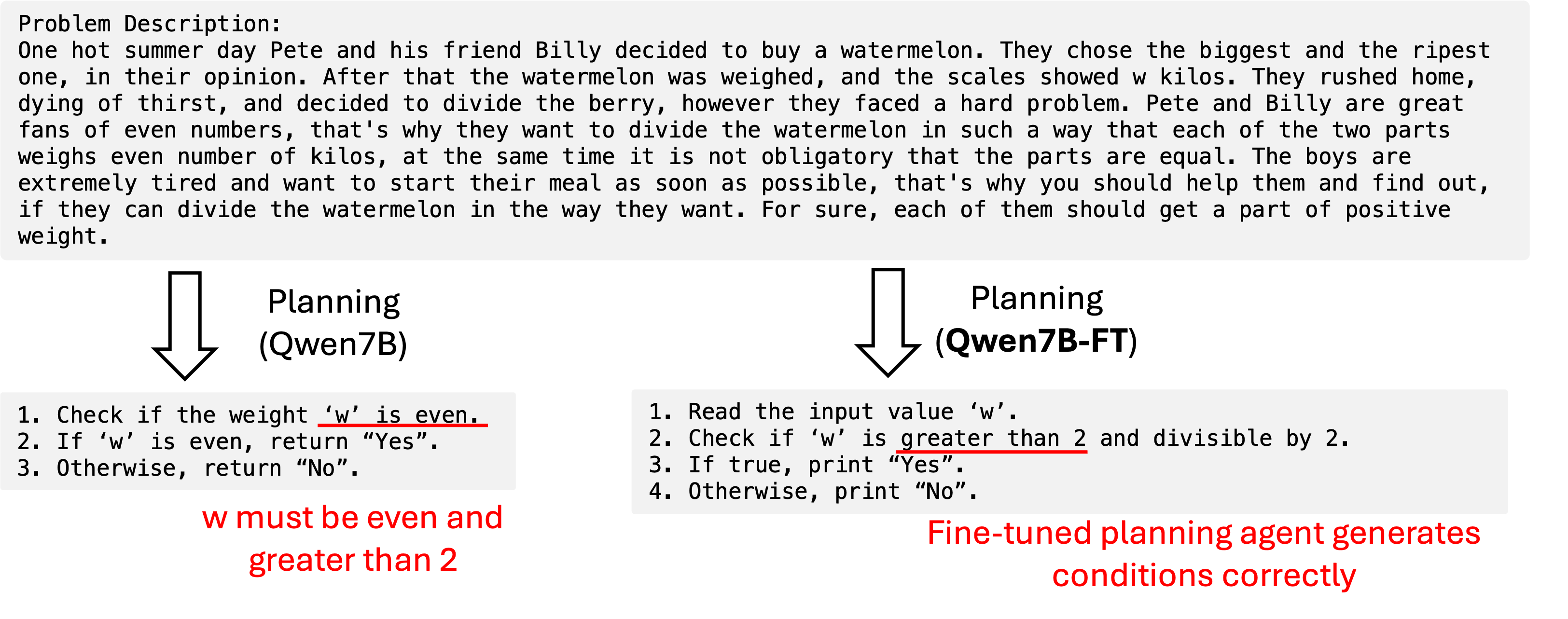}
\caption{Improvement in conditional logic by the planning agent after fine-tuning. The initial code fails to account for the edge case (\texttt{w > 2}) and only checks for evenness. After fine-tuning, the planning agent correctly generates the full condition required by the problem.}
\label{fig:planning-example}
\end{figure*}

\begin{figure*}[t]
\centering
\includegraphics[width=\textwidth]{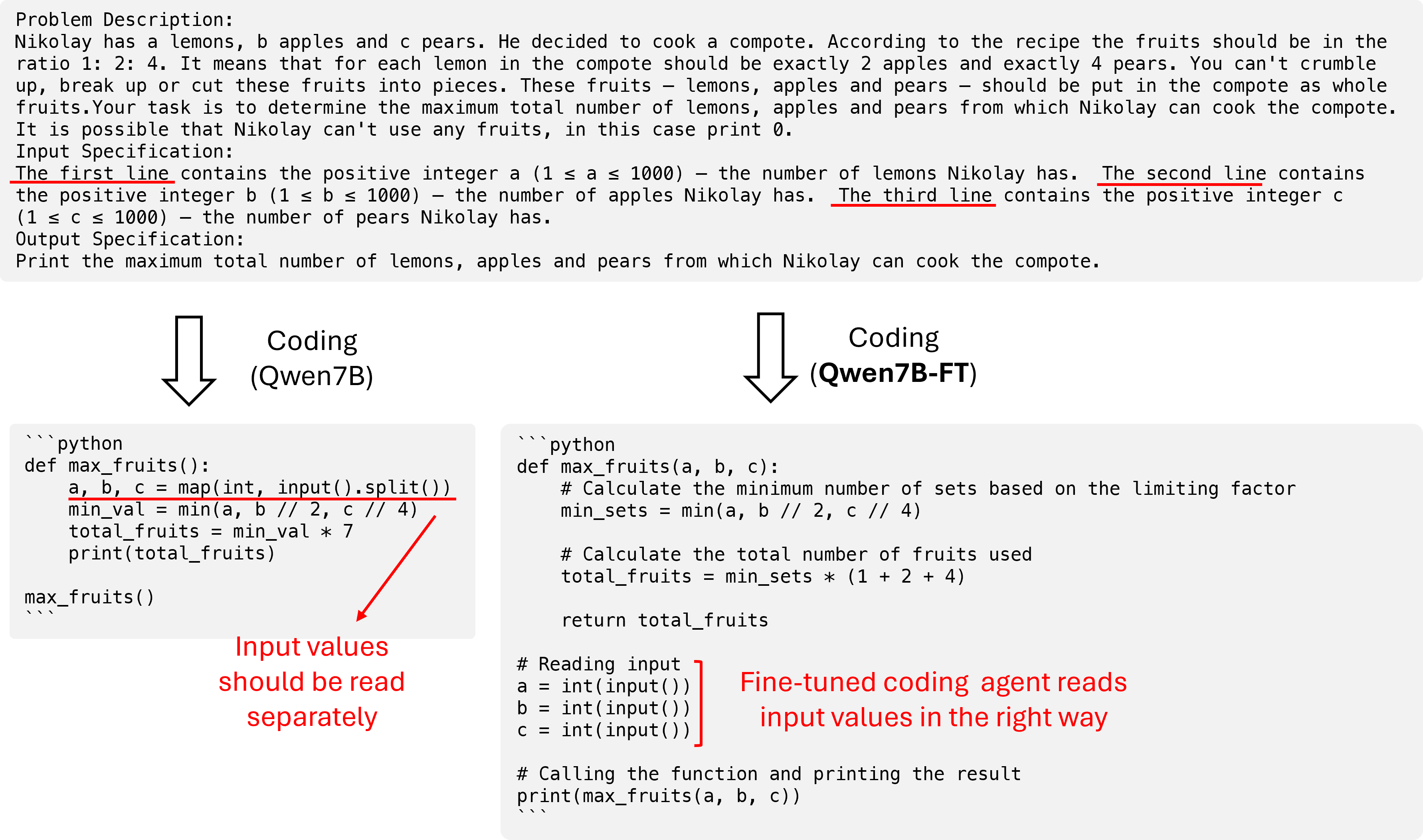}
\caption{Example of a coding error and its resolution after fine-tuning. The unfine-tuned agent incorrectly reads all input values from a single line, which violates the problem specification. After fine-tuning, the coding agent correctly processes line-separated inputs and computes the result accordingly.}
\label{fig:coding-example}
\end{figure*}

\begin{figure*}[t]
\centering
\includegraphics[width=\textwidth]{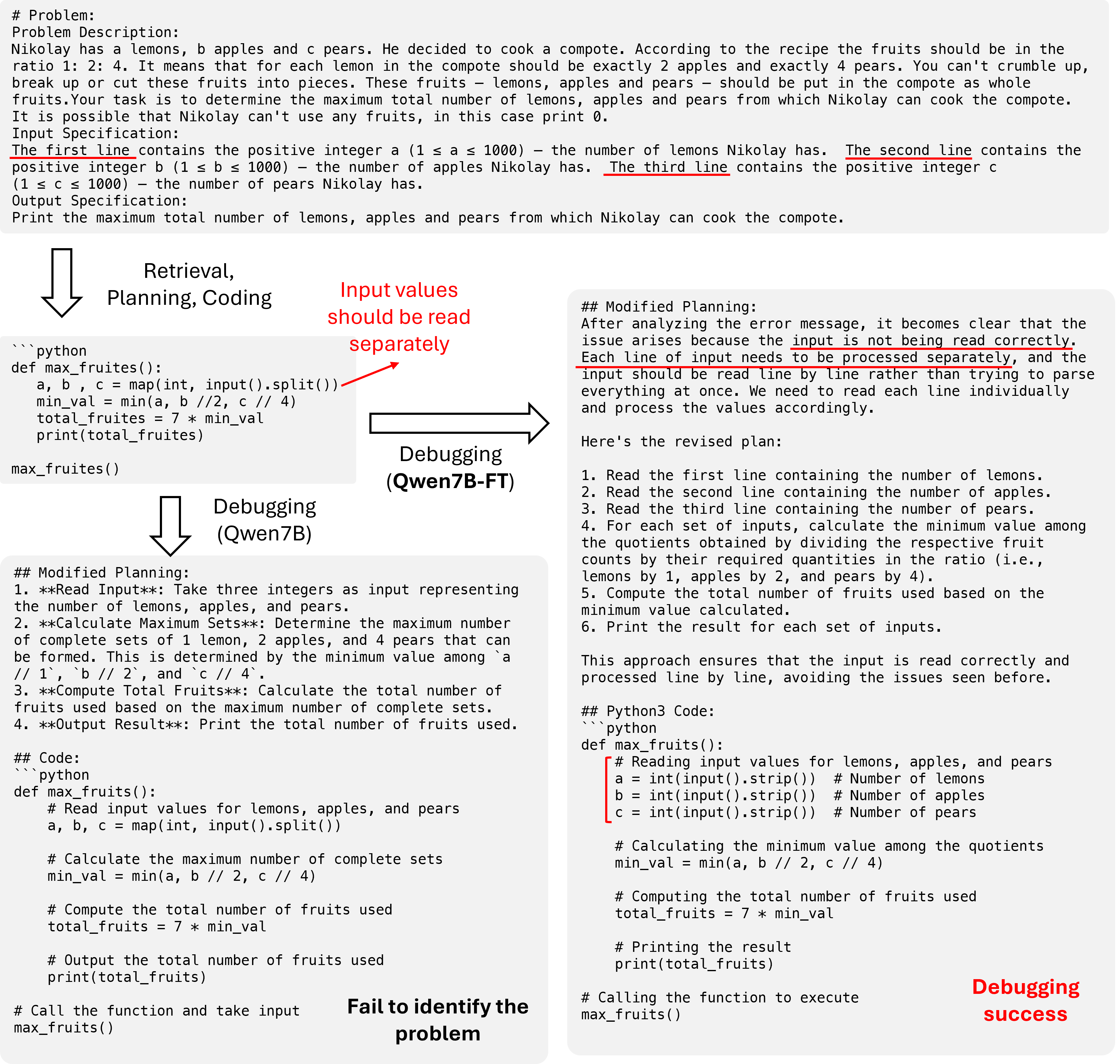}
\caption{Example of a debugging failure and recovery in the MapCoder pipeline. The initial plan and code fail to address the problem due to incorrect input parsing. After the debugging agent identifies the issue, the revised plan introduces line-by-line input reading, which successfully resolves the error.}
\label{fig:debugging-example}
\end{figure*}

Fig.~\ref{fig:retrieval-example},~\ref{fig:planning-example},~\ref{fig:coding-example}, and~\ref{fig:debugging-example} illustrate representative improvements observed in the 7B model for the retrieval, planning, coding, and debugging agents, respectively.

In the retrieval example (Fig.~\ref{fig:retrieval-example}), the pre-fine-tuned model incorrectly identifies the core algorithm as “Sliding Window” and includes unsupported tags like \texttt{<description>} within the \texttt{<algorithm>} block, failing to conform to the expected XML schema. After fine-tuning, the agent accurately identifies the algorithm (“Counting and Matching Pairs”) and outputs a well-structured XML response with all required tags properly closed.

In the planning example (Fig.~\ref{fig:planning-example}), the original plan fails to capture key logical conditions (e.g., that the input must be both even and greater than 2). After fine-tuning, the agent successfully generates a plan that handles these conditions explicitly and correctly guides the coding agent.

In the coding example (Fig.~\ref{fig:coding-example}), the unfine-tuned model incorrectly processes multiple input values on a single line, violating the problem’s input specification. Fine-tuning enables the agent to read inputs line by line, aligning its behavior with the expected input format and improving functional correctness.

Finally, the debugging example (Fig.~\ref{fig:debugging-example}) demonstrates how the unfine-tuned model fails to fix a bug caused by improper input parsing. The fine-tuned debugging agent correctly diagnoses the issue and proposes a revised plan and code that successfully passes all test cases.


\onecolumn
\section{Trajectory Example}
\label{sec:trajectory}

These are example trajectories used for fine-tuning each agent. Each consists of an input–output pair, and the examples are drawn from different problems.

\tcbset{
  myboxstyle/.style={
    colback=teal!5!white,       
    colframe=teal!60!black,     
    coltitle=black,             
    colbacktitle=teal!60!white, 
    fonttitle=\bfseries,        
    boxrule=0.6pt,              
    arc=2mm,                    
    left=3mm, right=3mm, top=2mm, bottom=2mm,
    title style={top color=white,bottom color=teal!15!white},
  }
}

\subsection{Retrieval Agent}

\begin{tcolorbox}[myboxstyle, title={Example Trajectory of Retrieval (Input)}, width=\textwidth, breakable]
Given a problem, algorithm tags, and solution code, provide relevant problems then identify the algorithm behind it and also explain the tutorial of the algorithm.\\
\# Problem:\\
You are given several queries. Each query consists of three integers \(p\), \(q\) and \(b\). You need to answer whether the result of \(p/q\) in notation with base \(b\) is a finite fraction.\\
\\
A fraction in notation with base \(b\) is finite if it contains finite number of numerals after the decimal point. It is also possible that a fraction has zero numerals after the decimal point.\\
\\
-----Input-----\\
\\
The first line contains a single integer \(n\) (\(1 \le n \le 10^5\)) --- the number of queries.\\
\\
Next \(n\) lines contain queries, one per line. Each line contains three integers \(p\), \(q\), and \(b\) (\(0 \le p \le 10^{18}\), \(1 \le q \le 10^{18}\), \(2 \le b \le 10^{18}\)). All numbers are given in notation with base 10.\\
\\
-----Output-----\\
\\
For each question, in a separate line, print Finite if the fraction is finite and Infinite otherwise.\\
\\
-----Examples-----\\
Input\\
2\\
6 12 10\\
4 3 10\\
\\
Output\\
Finite\\
Infinite\\
\\
Input\\
4\\
1 1 2\\
9 36 2\\
4 12 3\\
3 5 4\\
\\
Output\\
Finite\\
Finite\\
Finite\\
Infinite\\
\\
-----Note-----\\
\\
\(\frac{6}{12} = \frac{1}{2} = 0.5_{10}\)\\
\\
\(\frac{4}{3} = 1.(3)_{10}\)\\
\\
\(\frac{9}{36} = \frac{1}{4} = 0.01_2\)\\
\\
\(\frac{4}{12} = \frac{1}{3} = 0.1_3\)\\
\\
Sample Input Format:\\
2\\
6 12 10\\
4 3 10\\
\\
Sample Output Format:\\
Finite\\
Infinite\\
\\
-------\\
Important: You must follow the input output format. Input should be taken from standard input and output should be given to standard output.\\
Note: If you are writing a function then after the function definition take input from using input() function, call the function with specified parameters and finally print the output of the function.\\
\\
\# Exemplars:\\
Recall three (03) relevant and distinct problems (different from problem mentioned above). For each problem,\\
1. describe it\\
2. generate Python3 code step by step to solve that problem\\
3. finally generate a planning to solve that problem\\
\\
\# Algorithm:\\
\\
----------------\\
Important:\\
Your response must follow the following xml format-\\
\\
<root>\\
<problem>\\
\# Recall three (03) relevant and distinct problems (different from problem mentioned above). Write each problem in the following format.\\
<description>\\
\# Describe the problem.\\
</description>\\
<code>\\
\# Let's think step by step to solve this problem in Python3 programming language.\\
</code>\\
<planning>\\
\# Planning to solve this problem.\\
</planning>\\
</problem>\\
\\
\# similarly add more problems here...\\
\\
<algorithm>\\
\# Identify the algorithm (Brute-force, Dynamic Programming, Divide-and-conquer, Greedy, Backtracking, Recursive, Binary search, and so on) that needs to be used to solve the original problem.\\
\# Write a useful tutorial about the above mentioned algorithms. Provide a high level generic tutorial for solving this types of problem. Do not generate code.\\
</algorithm>\\
</root>
\end{tcolorbox}

\begin{tcolorbox}[myboxstyle, title={Example Trajectory of Retrieval (Output)}, width=\textwidth, breakable]
<root>\\
<problem>\\
<description>\\
Given a fraction \(p/q\), determine if it can be represented as a finite decimal in base \(b\). The fraction is finite in base \(b\) if the denominator \(q\) (after simplifying the fraction) has no prime factors other than those of the base \(b\).\\
</description>\\
<code>\\
import math\\
\\
def solve():\\
\ \ \ \ n = int(input())\\
\ \ \ \ for \_ in range(n):\\
\ \ \ \ \ \ \ \ p, q, b = map(int, input().split())\\
\ \ \ \ \ \ \ \ \# Simplify the fraction p/q\\
\ \ \ \ \ \ \ \ g = math.gcd(p, q)\\
\ \ \ \ \ \ \ \ q\_simplified = q // g\\
\ \ \ \ \ \ \ \ \# Remove all common factors between q\_simplified and b\\
\ \ \ \ \ \ \ \ while True:\\
\ \ \ \ \ \ \ \ \ \ \ \ g = math.gcd(q\_simplified, b)\\
\ \ \ \ \ \ \ \ \ \ \ \ if g == 1:\\
\ \ \ \ \ \ \ \ \ \ \ \ \ \ \ \ break\\
\ \ \ \ \ \ \ \ \ \ \ \ q\_simplified //= g\\
\ \ \ \ \ \ \ \ if q\_simplified == 1:\\
\ \ \ \ \ \ \ \ \ \ \ \ print("Finite")\\
\ \ \ \ \ \ \ \ else:\\
\ \ \ \ \ \ \ \ \ \ \ \ print("Infinite")\\
\\
solve()\\
</code>\\
<planning>\\
1. Read the number of queries.\\
2. For each query, read \(p\), \(q\), and \(b\).\\
3. Simplify the fraction \(p/q\) by dividing both numerator and denominator by their greatest common divisor (GCD).\\
4. Repeatedly divide the simplified denominator by the GCD of the denominator and the base \(b\) until no more common factors exist.\\
5. If the resulting denominator is 1, the fraction is finite in base \(b\); otherwise, it is infinite.\\
6. Print "Finite" or "Infinite" based on the result.\\
</planning>\\
</problem>\\
(Other example problems...)\\
<algorithm>\\
The algorithm used to solve the original problem is based on number theory, specifically the properties of fractions and their representations in different bases. The key insight is that a fraction \(p/q\) has a finite representation in base \(b\) if and only if the denominator \(q\) (after simplifying the fraction) has no prime factors other than those of the base \(b\).\\
\\
Tutorial:\\
1. Simplify the fraction \(p/q\) by dividing both numerator and denominator by their GCD.\\
2. Remove all common prime factors between the simplified denominator and the base \(b\) by repeatedly dividing the denominator by the GCD of the denominator and the base.\\
3. If the resulting denominator is 1, the fraction is finite in base \(b\); otherwise, it is infinite.\\
\\
This approach efficiently checks the finiteness of the fraction by leveraging GCD operations to eliminate common factors, ensuring optimal performance even for large inputs.\\
</algorithm>\\
</root>
\end{tcolorbox}

\subsection{Planning Agent}

\begin{tcolorbox}[myboxstyle, title={Example Trajectory of Planning (Input)}, width=\textwidth, breakable]
Given a competitive programming problem generate a concrete planning to solve the problem.\\
\# Problem:\\

Given a string, reverse each word in the string while keeping the order of the words the same. For example, "Hello World" becomes "olleH dlroW".\\

\# Planning:\\

1. Split the input string into individual words using the `split()` method.\\
2. Iterate over each word and reverse it using slicing (`[::-1]`).\\
3. Join the reversed words back into a single string with spaces separating them.\\
4. Print the result.\\

\#\# Relevant Algorithm to solve the next problem:\\

The algorithm used to solve the original problem (extracting the first character of each word in a string) is a **String Manipulation** technique, specifically involving splitting the string into words and processing each word individually. \\

\#\#\# Tutorial: String Manipulation for Word Processing\\
1. **Splitting the String**: Use the `split()` method to divide the input string into a list of words based on spaces.\\
2. **Processing Each Word**: Iterate over the list of words and extract the first character of each word.\\
3. **Combining Results**: Join the extracted characters into a single string to form the final output.\\

This approach is efficient for problems where you need to process individual words in a string. It leverages basic string operations and does not require complex algorithms like dynamic programming or backtracking. The key steps are splitting, iterating, and combining, which are fundamental in string manipulation tasks.\\

\#\# Problem to be solved:\\
In this exercise, a string is passed to a method and a new string has to be returned with the first character of each word in the string.\\

For example:\\
```\\
"This Is A Test" ==> "TIAT"\\
```\\\\
Sample Input Format:\\
sees eyes xray yoat\\
Sample Output Format:\\
sexy\\

-------\\
Important: You must follow the input output format. Input should be taken from standard input and output should be given to standard output.\\
Note: If you are writing a function then after the function definition take input from using `input()` function, call the function with specified parameters and finally print the output of the function.\\
\#\# Sample Test cases: \\ 
Input:\\
sees eyes xray yoat\\
Expected output:\\
sexy\\
Input:\\
brown eyes are nice\\
Expected output:\\
bean\\

\#\# Planning:\\

----------------\\
Important: You should give only the planning to solve the problem. Do not add extra explanation or words.\\
\end{tcolorbox}

\begin{tcolorbox}[myboxstyle, title={Example Trajectory of Planning (Output)}, width=\textwidth, breakable]
1. Split the input string into individual words using the `split()` method.\\
2. Iterate over each word and extract the first character.\\
3. Join the extracted characters into a single string.\\
4. Print the result.
\end{tcolorbox}

\subsection{Coding Agent}
\begin{tcolorbox}[myboxstyle, title={Example Trajectory of Coding (Input)}, width=\textwidth, breakable]
Given a competitive programming problem generate Python3 code to solve the problem.\\
\#\# Relevant Algorithm to solve the next problem:\\

The algorithm used in the original problem is **String Slicing**. \\

\#\#\# Tutorial: String Slicing\\
String slicing is a technique to extract parts of a string by specifying start, end, and step indices. The general syntax is \texttt{string[start:end:step]}, where:\\
- \texttt{start} is the index where the slice begins (inclusive).\\
- \texttt{end} is the index where the slice ends (exclusive).\\
- \texttt{step} determines the stride (e.g., 1 for forward, -1 for reverse).\\

\#\#\#\# Key Points:\\
1. **Indices**: Python uses zero-based indexing. Negative indices count from the end (\(-1\) is the last character).\\
2. **Defaults**: If omitted, \texttt{start} defaults to 0, \texttt{end} to \texttt{len(string)}, and \texttt{step} to 1.\\
3. **Applications**:\\
   - Extracting substrings (\texttt{s[1:4]}).\\
   - Reversing strings (\texttt{s[::-1]}).\\
   - Removing characters (\texttt{s[1:-1]} removes first and last characters).\\

\#\#\#\# High-Level Approach:\\
1. Identify the substring you need by determining the start and end indices.\\
2. Use slicing to extract the substring.\\
3. Handle edge cases (e.g., empty strings, out-of-bound indices).\\

This technique is efficient (\(O(k)\) for a slice of length \(k\)) and widely used in string manipulation tasks.\\

\#\# Problem to be solved:\\
It's pretty straightforward. Your goal is to create a function that removes the first and last characters of a string. You're given one parameter, the original string. You don't have to worry with strings with less than two characters.\\

Sample Input Format:\\
eloquent\\
Sample Output Format:\\
loquen\\

-------\\
Important: You must follow the input output format. Input should be taken from standard input and output should be given to standard output.\\
Note: If you are writing a function then after the function definition take input from using \texttt{input()} function, call the function with specified parameters and finally print the output of the function.\\
\#\# Planning:\\
1. Define a function \texttt{remove\_first\_last\_char} that takes a string \texttt{s} as input.  \\
2. Use string slicing to exclude the first and last characters (\texttt{s[1:-1]}).  \\
3. Take input from the user, call the function, and print the result.\\
\#\# Sample Test cases: \\ 
Input:\\
eloquent\\
Expected output:\\
loquen\\
Input:\\
country\\
Expected output:\\
ountr\\\\
\#\# Let's think step by step.\\\\
----------------\\
Important:\\

\#\# Your response must contain only the Python3 code to solve this problem. Do not add extra explanation or words.\\
\end{tcolorbox}

\begin{tcolorbox}[myboxstyle, title={Example Trajectory of Coding (Output)}, width=\textwidth, breakable]
```python
\begin{verbatim}
def remove_first_last_char(s):    
  return s[1:-1]
input_str = input()
print(remove_first_last_char(input_str))
\end{verbatim}
```
\end{tcolorbox}

\subsection{Debugging Agent}

\begin{tcolorbox}[myboxstyle, title={Example Trajectory of Debugging (Input)}, width=\textwidth, breakable]
Given a competitive programming problem you have generated Python3 code to solve the problem. But the generated code can not pass sample test cases. Improve your code to solve the problem correctly.\\
\#\# Relevant Algorithm to solve the next problem:\\

The algorithm used to solve the original problem of breaking up camel casing involves a straightforward approach to identifying capital letters within a string and inserting a space before them. This can be categorized under the Brute-force method since it directly iterates through the string and applies a simple condition to achieve the desired outcome.\\

**Tutorial on Brute-force Algorithms:**\\
Brute-force algorithms involve trying every possible solution until the correct one is found. They are often the simplest and most direct way to solve a problem but may not always be the most efficient, especially for larger datasets. The key aspects of implementing a brute-force algorithm include:\\
- Identifying all possible solutions.\\
- Checking each solution against the problem constraints.\\
- Selecting the first solution that meets all criteria.\\
In the context of the camel case problem, the brute-force approach directly iterates through the string, checks if a character is uppercase, and if so, inserts a space before it. This method is effective for the task at hand due to its simplicity and directness.\\

\#\# Problem to be solved:\\
Complete the solution so that the function will break up camel casing, using a space between words.\\

\#\#\# Example\\

``` \\
solution("camelCasing") == "camel Casing"\\
``` \\

Sample Input Format:\\
helloWorld\\
Sample Output Format:\\
hello World\\

-------\\
Important: You must follow the input output format. Input should be taken from standard input and output should be given to standard output.\\
Note: If you are writing a function then after the function definition take input from using \texttt{input()} function, call the function with specified parameters and finally print the output of the function.\\
\#\# Planning:\\
1. Identify the indices where a transition from lowercase to uppercase letter occurs.\\
2. Insert a space before each identified index.\\
3. Join the modified string segments.\\
4. Return or print the final result.\\
\#\# Code:\\
```\\
def solution(s):\\
\ \ \ \ return ''.join([' ' + c if c.isupper() else c for c in s])[1:]\\

s = input()\\
print(solution(s))\\
```\\

\#\# Test Report:\\
\#\# Tested passed:\\
No test cases passed.\\

\#\# Tests failed:\\
Input:\\
helloWorld\\
Expected Output:\\
hello World\\
Your Output:\\
ello World\\

Input:\\
camelCase\\
Expected Output:\\
camel Case\\
Your Output:\\
amel Case\\

\#\# Modified Planning:\\
\#\# Let's think step by step to modify Python3 Code for solving this problem.\\

----------------\\
Important:\\

\#\# Note: Strictly follow the input and output format. The input should be taken from Standard input and output should be given to standard output. If you are writing a function then after the function definition take input using \texttt{input()} function then call the function with specified parameters and finally print the output of the function. Do not add extra print statement otherwise it will failed the test cases.\\
\#\# Your response must contain the modified planning and then the Python3 code inside ``` block to solve this problem.\\
\end{tcolorbox}

\begin{tcolorbox}[myboxstyle, title={Example Trajectory of Debugging (Output)}, width=\textwidth, breakable]
\#\# Modified Planning:\\
1. **Identify Uppercase Letters**: Iterate through each character in the input string to check if it is uppercase.\\
2. **Insert Space Before Uppercase Letters**: For each uppercase character found, insert a space before it.\\
3. **Construct Result String**: Combine all characters, including the inserted spaces, to form the final result string.\\
4. **Handle Edge Cases**: Ensure that the first character of the string is not modified if it is uppercase (no space should be inserted before the first character).\\

The issue with the original code was that it was slicing the string from the second character (\texttt{[1:]}), which caused the first character to be omitted. This needs to be fixed by ensuring the first character is included in the result.\\

\#\# Modified Python3 Code:\\
```python\\
def solution(s):\\
\ \ \ \ result = []\\
\ \ \ \ for c in s:\\
\ \ \ \ \ \ \ \ if c.isupper():\\
\ \ \ \ \ \ \ \ \ \ \ \ result.append(' ' + c)\\
\ \ \ \ \ \ \ \ else:\\
\ \ \ \ \ \ \ \ \ \ \ \ result.append(c)\\
\ \ \ \ return ''.join(result)\\

s = input()\\
print(solution(s))\\
```\\
\end{tcolorbox}

\end{document}